\documentclass{article} 

\usepackage[acronym,nowarn,section,nogroupskip,nonumberlist]{glossaries}

\glsdisablehyper{}
\newcommand{\acaps}[1]{{\scshape #1}}
\newacronym[\glslongpluralkey={Conditional Neural Processes}]{cnp}{\acaps{cnp}}{Conditional Neural Process}

\usepackage{amsmath, amssymb, amsthm}
\usepackage{mathtools}
\usepackage{bbm}
\usepackage{dsfont}

\newcommand{\bH}{\mathbf{H}}

\newcommand{\bsm}{\boldsymbol{m}}

\newcommand{\bsw}{\boldsymbol{w}}

\newcommand{\calB}{{\mathcal{B}}}

\newcommand{\calD}{{\mathcal{D}}}

\newcommand{\calL}{{\mathcal{L}}}

\newcommand{\bbR}{\mathbb{R}}

\theoremstyle{plain}

\theoremstyle{definition}

\theoremstyle{remark}

\def\[#1\]{\begin{equation}\begin{aligned}#1\end{aligned}\end{equation}}

\usepackage[usenames,dvipsnames]{xcolor}

\usepackage{pifont}
\newcommand{\cmark}{\ding{51}}%
\newcommand{\xmark}{\ding{55}}%

\usepackage{graphicx, wrapfig, float}
\usepackage{stfloats}
\usepackage[labelfont=bf]{caption}
\usepackage[format=hang]{subcaption}

\usepackage{booktabs, array, multirow, makecell, tabularx}
\usepackage[usestackEOL]{stackengine}
\newcommand{\spm}[1]{\scriptstyle{\pm\text{#1}}}

\usepackage[colorlinks, linktoc=all,pagebackref=true]{hyperref}
\usepackage[all]{hypcap}
\hypersetup{citecolor=MidnightBlue}
\hypersetup{linkcolor=RawSienna}
\hypersetup{urlcolor=MidnightBlue}
\usepackage[nameinlink,capitalise, noabbrev]{cleveref}
\creflabelformat{equation}{#2\textup{#1}#3}  
\crefname{section}{\S}{\S\S}

\newsavebox\CBox 
\def\BF#1{\sbox\CBox{#1}\resizebox{\wd\CBox}{\ht\CBox}{\textbf{#1}}}

\def\CP#1{\scriptsize{\color{OliveGreen}(\texttt{+}#1)}}

\usepackage{iclr2024_conference,times}

\title{Sparse Weight Averaging with Multiple Particles for Iterative Magnitude Pruning}
\author{%
    Moonseok Choi$^{*1}$ \quad Hyungi Lee$^{*1}$ \quad Giung Nam$^{*1}$ \quad Juho Lee$^{1,2}$ \\
    $^1$KAIST AI \quad $^2$AITRICS\\
    \texttt{\{ms.choi, lhk2708, giung, juholee\}@kaist.ac.kr}
}

\iclrfinalcopy

\newcommand\blfootnote[1]{
    \begingroup
    \renewcommand\thefootnote{}\footnote{#1}
    \addtocounter{footnote}{-1}
    \endgroup
}

\usepackage{algorithm}
\usepackage{algpseudocode}

\begin{document}
\blfootnote{$^*$ Equal contribution}

\maketitle

\begin{abstract}
Given the ever-increasing size of modern neural networks, the significance of sparse architectures has surged due to their accelerated inference speeds and minimal memory demands. 
When it comes to global pruning techniques, Iterative Magnitude Pruning (IMP) still stands as a state-of-the-art algorithm despite its simple nature, particularly in extremely sparse regimes.
In light of the recent finding that the two successive matching IMP solutions are linearly connected without a loss barrier, we propose Sparse Weight Averaging with Multiple Particles (SWAMP), a straightforward modification of IMP that achieves performance comparable to an ensemble of two IMP solutions.
For every iteration, we concurrently train multiple sparse models, referred to as particles, using different batch orders yet the same matching ticket, and then weight average such models to produce a single mask. 
We demonstrate that our method consistently outperforms existing baselines across different sparsities through extensive experiments on various data and neural network structures.
\end{abstract}


\section{Introduction}
\label{main:sec:introduction}





Deep neural networks are often highly over-parameterized, and the majority of their parameters can be pruned without sacrificing model performance~\citep{lecun1989optimal}. The \textit{lottery ticket hypothesis} proposed by \citet{frankle2019lottery} suggests that there exists a sparse subnetwork at initialization that can be trained to achieve the same level of performance as the original dense network. Such \textit{matching} subnetworks can be found via Iterative Magnitude Pruning (IMP) with rewinding~\citep{frankle2020linear}, which involves the following three steps: (i) training the network for a certain number of iterations, (ii) pruning weights with the smallest magnitudes, and (iii) rewinding the weights back to an early iteration while fixing the pruned weights to zero. This procedure is repeated for several rounds, and the final rewound subnetwork corresponds to the \textit{matching ticket} that can achieve the performance of the full network. Despite its simplicity, IMP offers the state-of-the-art performance as to finding a sparse mask, especially for extreme sparsity regimes~\citep{renda2020comparing}.

The success of IMP is indeed counter-intuitive considering its simplicity. In this regard, \citet{frankle2020linear} revealed an underlying connection between the lottery ticket hypothesis and linear mode connectivity, indicating that the effectiveness of IMP is reliant upon its stability to stochastic optimization; IMP solutions reside in the same basin of attraction in the loss landscape. Delving deeper into the subject matter, \citet{paul2023unmasking} found that linear mode connectivity also exists between successive IMP solutions with different sparsity levels. More precisely, they concluded that IMP fails to find a \textit{matching} subnetwork if the solutions from consecutive rounds are disconnected and further highlighted the significance of both the pruning ratio and the rewinding iteration to retain the connectivity between IMP solutions.

Inspired by the connection between IMP and linear mode connectivity, we expand the understanding to the \textit{loss landscape perspective}. Analyzing the loss landscape of deep neural networks is an effective tool that is widely employed to study mode connectivity~\citep{draxler2018essentially,garipov2018loss,fort2019large,benton2021loss}, and it also motivates us to find solutions located at the flat region of the loss landscape to enhance generalization~\citep{chaudhari2017entropysgd,izmailov2018averaging,foret2021sharpnessaware}. Notably, both fields share a common objective of identifying ``good'' subspaces characterized by low loss value, and this objective aligns with the ultimate goal of neural network pruning - to identify ``matching'' sparse subspaces within a given dense parameter space.

In this paper, we study how IMP can benefit from the multiple models connected in loss surfaces. Our contributions are summarized as follows: 

\begin{itemize}
    \item We first empirically demonstrate that multiple models trained with different SGD noise yet from the same matching ticket can be weight-averaged, i.e., there exists no loss barrier within the convex hull of the model weights. We further show that taking an average of the particles leads to flat minima, which exhibit superior generalization performance compared to each individual particle.

    \item Building upon prior observations, we propose a novel iterative pruning technique, Sparse Weight Averaging with Multiple Particles (SWAMP), tailored specifically for IMP. We verify that SWAMP preserves the linear connectivity of successive solutions, which is a crucial feature contributing to the success of IMP.

    \item Through extensive experiments,
     we provide empirical evidence that supports the superiority of the proposed SWAMP algorithm over other baselines.
\end{itemize}

\section{Backgrounds}
\label{main:sec:backgrounds}

\subsection{Neural network pruning as constrained optimization}

Conventional training of neural networks aims to find an optimal neural network parameter $\bsw\in\bbR^D$ that minimizes a given loss function $\calL : \bbR^D \rightarrow \bbR$ for a given training dataset $\calD$. Such optimization typically employs the Stochastic Gradient Descent~\citep[SGD;][]{robbins1951stochastic} methods, which we denote as $\bsw_{T} \gets \operatorname{SGD}_{0 \rightarrow T}(\bsw_{0}, \xi, \calD)$ throughout the paper. Here, $\bsw_T$ denotes the solution obtained by performing SGD with a randomness of $\xi$ (e.g., mini-batch ordering) over $T$ iterations, starting from the initial weight $\bsw_0$. On the other hand, neural network pruning is the process of obtaining a sparse neural network with a desired sparsity level $\kappa \in [0, 1)$ from the original dense neural network. The goal is now to find an \textit{optimal sparse solution} $\bsw=\bsw^\ast\circ\bsm^\ast$ subject to the constraint that the number of non-zero elements in the mask $\smash{\bsm^\ast\in[0,1]^D}$ satisfies $\smash{\lVert\bsm^\ast\rVert_0 \leq D(1-\kappa)}$.

\subsection{Iterative magnitude pruning with rewinding}

Iterative Magnitude Pruning~\citep[IMP;][]{frankle2019lottery} is an iterative pruning method that is both straightforward and highly effective. Each cycle of IMP involves the following three steps: \textit{(i) Train} - a network parameter $\bsw_c$ at the $c^\text{th}$ cycle is trained until it reaches convergence. \textit{(ii) Prune} - a mask $\bsm$ is created by setting the smallest weights to zero based on a predefined pruning ratio of $\alpha$. \textit{(iii) Reset} - the weights are then reverted back to their initial values before the next cycle begins. This \textit{train-prune-reset} cycle is repeated until the desired level of sparsity is achieved.

However, in practical scenarios, the original version of IMP suffers from rapid performance degradation as sparsity increases and fails to match the performance of the original dense solution. To address this issue, the concept of \textit{rewinding} is introduced~\citep{frankle2020linear,renda2020comparing}. Rather than \textit{resetting} the unpruned weights to their initial values, the weights are \textit{rewound} to an early training point - the \textit{matching ticket}. The matching ticket is simply the weights obtained after training for a few iterations. Refer to~\cref{app:sec:additional_experiments:algorithms} for more details on IMP algorithms.

\subsection{Linear connectivity of neural network weights}

Consider a one-dimensional path denoted as $\smash{P : [0, 1] \rightarrow \bbR^D}$, connecting two neural network weights $\smash{\bsw^{(0)}}$ and $\smash{\bsw^{(1)}}$ in a $D$-dimensional space, where the starting and end points are $\smash{P(0) = \bsw^{(0)}}$ and $\smash{P(1) = \bsw^{(1)}}$, respectively. In a simplified sense, we can say that there is a \textit{connectivity} between $\smash{\bsw^{(0)}}$ and $\smash{\bsw^{(1)}}$ if the condition $\smash{\sup_{\lambda\in[0,1]} \calL(P(\lambda)) \leq \max{\{\calL(P(0)), \calL(P(1))\}} + \epsilon}$ holds, where $\epsilon$ is a small margin value. While recent advances in deep learning have revealed the existence of non-linear paths between local minima obtained through stochastic optimization~\citep{draxler2018essentially,garipov2018loss}, it is still not straightforward to establish linear connectivity (i.e., connectivity with a linear connector $\smash{P(\lambda)=(1-\lambda)\bsw^{(0)} + \lambda\bsw^{(1)}}$) for modern deep neural networks~\citep{lakshminarayanan2017simple,fort2019large,fort2020deep}.

\section{Sparse Weight Averaging with Multiple Particles (SWAMP)}
\label{main:sec:methods}

\begin{figure}[t]
\centering
\includegraphics[width=\linewidth]{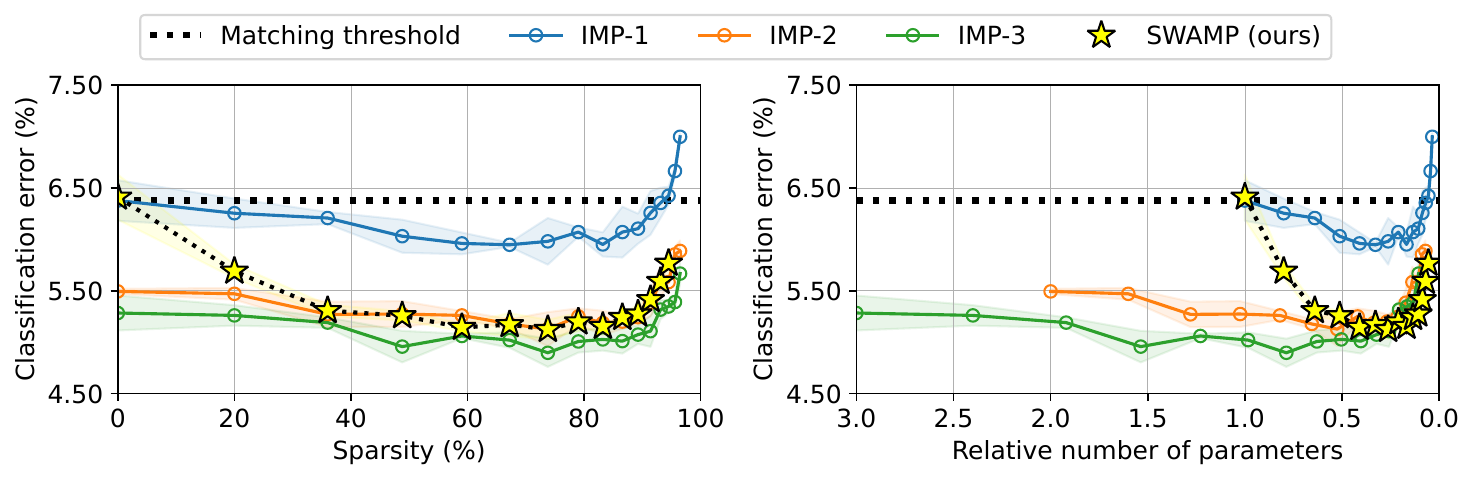}
\caption{Classification error as a function of the sparsity (left) and the relative number of parameters (right). Our proposed SWAMP achieves remarkable performance comparable to an ensemble of IMP solutions, where IMP-$n$ indicates the ensemble of $n$ IMP solutions, while maintaining the same inference cost as a single model. Notably, SWAMP demonstrates \textit{matching} performance even at extremely sparse levels, unlike IMP. The results are presented for WRN-28-2 on CIFAR-10, and we refer readers to~\cref{app:sec:additional_experiments:sparsity} for the same plot for CIFAR-100, as well as VGG architectures.}
\label{figure/sparsity}
\end{figure}

\subsection{IMP: a loss landscape perspective}
\label{main:subsec:imp}

\citet{frankle2020linear} demonstrated that the matching ticket has a significant impact on the \textit{stability} of neural networks to SGD noise $\xi$. Even when two networks are trained with the same random initialization $\bsw_0$, the different SGD noise $\xi^{(1)}$, $\xi^{(2)}$ disrupts the linear connectivity between the solutions obtained through SGD, i.e., there is no linear connector between
\[
\bsw_T^{(1)} \gets \operatorname{SGD}_{0 \rightarrow T}(\bsw_0, \xi^{(1)}, \calD) \text{ and }
\bsw_T^{(2)} \gets \operatorname{SGD}_{0 \rightarrow T}(\bsw_0, \xi^{(2)}, \calD),
\]
and thus the optimization is rendered unstable to SGD noise. They further empirically confirmed that sparse solutions obtained through IMP are matching \textit{if and only if} they are stable to SGD noise, and diagnosed this instability as a failure case of the original IMP algorithm~\citep{frankle2019lottery}. A simple treatment to ensure the stability is sharing the early phase of the optimization trajectory. In other words, there exists a linear connector between
\[
\bsw_T^{(1)} \gets \operatorname{SGD}_{T_0 \rightarrow T}(\bsw_{T_0}, \xi^{(1)}, \calD) \text{ and }
\bsw_T^{(2)} \gets \operatorname{SGD}_{T_0 \rightarrow T}(\bsw_{T_0}, \xi^{(2)}, \calD),
\]
when SGD runs are started from the same initialization of $\bsw_{T_0} \gets \operatorname{SGD}_{0 \rightarrow T_0}(\bsw_{0},\xi,\calD)$. Furthermore, \citet{paul2023unmasking} demonstrated linear connectivity between two consecutive IMP solutions with different sparsity levels and identified it as a crucial factor for the success of IMP.

Nevertheless, the question of whether a low-loss subspace is formed by the convex hull of three or more solutions remains uncertain, despite the presence of linear connectivity between each pair of solutions. If it becomes feasible to construct a low-loss volume subspace using IMP solutions, it could potentially yield a more effective solution with improved generalization at the midpoint of this subspace~\citep{wortsman2021learning}.

\subsection{SWAMP: an algorithm}
\label{main:subsec:swamp_algorithm}

\begin{algorithm}[t]
\caption{Iterative Magnitude Pruning {\color{RoyalBlue}with SWAMP (ours)}}
\label{table/algorithm_swamp}
\begin{algorithmic}[1]
\Require Neural network parameter $\bsw$, pruning mask $\bsm$, training dataset $\calD$, the number of cycles for iterative magnitude pruning $C$, the number of iterations for each cycle $T$, pruning ratio $\alpha$, SGD noise $\xi$, the number of iteration for matching ticket $T_0$, and the number of particles $N$.
\Ensure Sparse solution $\overline{\bsw}_{c,T}$.
\item[]
\State Randomly initialize $\bsw_{0,0}$ and set mask $\bsm \gets \boldsymbol{1}$. \Comment{Starts from random dense weights.}
\State Train $\bsw_{0,T_0} \gets \textsc{SGD}_{0 \rightarrow T_0}(\bsw_{0,0}\circ\bsm, \xi_0, \calD)$. \Comment{Gets a matching ticket from the initialization.}
\For{$c\in\{1,\dots,C\}$}
    \color{RoyalBlue}
    \For{$n\in\{1,\dots,N\}$}
        \State Rewind $\bsw_{c,0}^{(n)} \gets \bsw_{0,T_0} \circ \bsm$. \Comment{Starts from the matching ticket.}
        \State Train $\bsw_{c,T}^{(n)} \gets \textsc{SWA}_{0 \rightarrow T}(\bsw_{c,0}^{(n)}, \xi_c^{(n)}, \calD)$. \Comment{Averages weights over trajectory.}
    \EndFor
    \State Averaging $\overline{\bsw}_{c,T} \gets \sum_{n=1}^{N} \bsw_{c,T}^{(n)} / N$. \Comment{Averages weights of particles.}
    \color{black}
    \State Prune $\bsm \gets \operatorname{Prune}({\color{RoyalBlue}\overline{\bsw}_{c,T}}, \alpha)$. \Comment{Updates the mask based on magnitudes.}
\EndFor
\end{algorithmic}
\end{algorithm}


Inspired by the stability analysis of the matching ticket presented in \cref{main:subsec:imp}, we propose Sparse Weight Averaging with Multiple Particles (SWAMP) as a tailored sparse weight averaging technique for IMP. The detailed algorithm is presented in \cref{table/algorithm_swamp}.

SWAMP differs from vanilla IMP in two main aspects. Firstly, we create multiple copies of the matching ticket (line 5; \cref{table/algorithm_swamp}) and train them simultaneously with different random seeds (line 6; \cref{table/algorithm_swamp}), whereas IMP employs a single particle. Secondly, we replace SGD training with Stochastic Weight Averaging~\citep[SWA;][]{izmailov2018averaging}, a method that constructs a moving average of parameters by periodically sampling a subset of the learning trajectory, and SWA enables us to accumulate virtually more particles throughout the training. We then average all particles before proceeding to the pruning step (line 8; \cref{table/algorithm_swamp}).

As illustrated in \cref{figure/sparsity}, our algorithm achieves superior performance, which is on par with that of an ensemble consisting of two sparse networks. This is quite remarkable considering that our solution achieves this level of performance while having significantly lower inference costs compared to the ensembling approach. Further ablation studies presented in \cref{main:subsec:ablation,app:sec:additional_experiments:ablation} also validate that both ingredients independently contribute to our algorithm, with each playing a crucial role in achieving superior performance.


\begin{figure}
\centering
\includegraphics[width=0.30\linewidth]{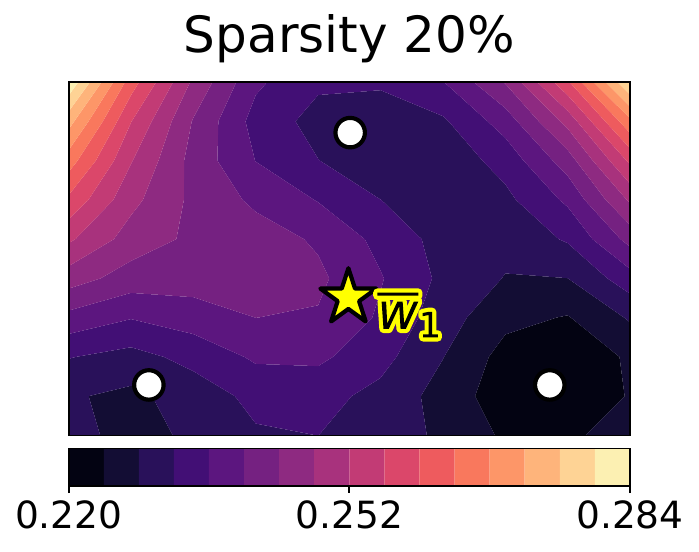}\hfill
\includegraphics[width=0.30\linewidth]{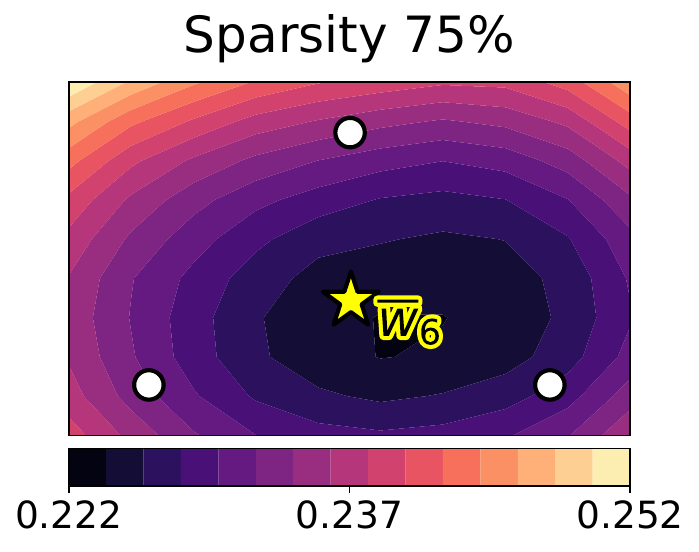}\hfill
\includegraphics[width=0.30\linewidth]{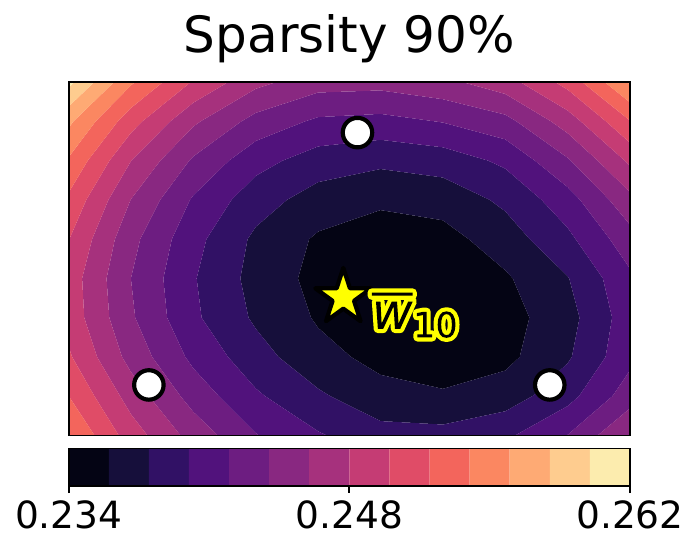}
\caption{Visualization of loss surfaces as a function of network weights in a two-dimensional subspace, spanned by three particles (marked as white circles). The averaged weight $\bsw_c$ (marked by a yellow star) is observed not to be positioned in the flat region of the surface during the earlier stages of IMP (left; Sparsity 20\%). However, as sparsity increases, the weight averaging technique effectively captures the flat region of the surface. The results are presented for WRN-28-2 on the test split of CIFAR-10, and we refer the reader to \cref{app:sec:additional_experiments:loss_surface} for the same plot for CIFAR-100.}
\label{figure/loss_surface}
\end{figure}


\begin{table}[t!]
\centering
\caption{Trace of Hessian $\text{Tr}(\bH)$ evaluated across training data. In most cases, SWAMP with multiple particles exhibits smaller trace value, i.e., finds flatter minima, compared to others. Reported values are averaged over three random seeds, and the best and second-best results are boldfaced and underlined, respectively. Refer to \cref{app:sec:additional_experiments:hessian} for the results on CIFAR-100.}
\label{table/main_hessian_trace}
\resizebox{0.95\linewidth}{!}{\begin{tabular}{clcccc}
    \toprule
    & & \multicolumn{4}{c}{Sparsity}\\
    \cmidrule(lr){3-6}
    & Training & 20\% & 50\% & 75\% & 90\%  \\
    \midrule
    \multirow{3}{*}{\Centerstack{CIFAR-10\\(WRN-28-2)}}	
        & SGD           & \phantom{}1784.64 $\spm{\phantom{}152.95}$ & \phantom{}1704.69 $\spm{\phantom{}142.57}$ & \phantom{}1719.49 $\spm{\phantom{}114.25}$ & \phantom{}1710.28 $\spm{\phantom{}185.84}$ \\ 
        & SWAMP ($N=1$) & \phantom{0}\underline{500.24} $\spm{\phantom{0}50.00}$ & \phantom{0}\underline{500.70} $\spm{\phantom{0}21.57}$ & \phantom{0}\underline{543.28} $\spm{\phantom{0}31.27}$ & \phantom{0}\underline{579.79} $\spm{\phantom{0}20.69}$ \\ 
        & SWAMP ($N=4$) & \phantom{0}{\bf{463.90}} $\spm{\phantom{0}34.84}$ & \phantom{0}{\bf{467.45}} $\spm{\phantom{0}35.76}$ & \phantom{0}{\bf{515.54}} $\spm{\phantom{0}20.23}$ & \phantom{0}{\bf{533.62}} $\spm{\phantom{00}4.27}$ \\ 
    \midrule
    \multirow{3}{*}{\Centerstack{CIFAR-10\\(VGG-13)}}
        & SGD           & \phantom{}1155.64 $\spm{\phantom{0}51.21}$ & \phantom{}1231.69 $\spm{\phantom{0}55.21}$ & \phantom{}1143.70 $\spm{\phantom{0}65.25}$ & \phantom{}1197.73 $\spm{\phantom{0}21.04}$ \\	
        & SWAMP ($N=1$) & \phantom{0}{\bf{427.92}} $\spm{\phantom{0}11.15}$ & \phantom{0}\underline{424.85} $\spm{\phantom{00}6.84}$ & \phantom{0}\underline{450.44} $\spm{\phantom{0}20.82}$ & \phantom{0}\underline{473.89} $\spm{\phantom{0}55.69}$ \\ 
        & SWAMP ($N=4$) & \phantom{0}\underline{432.45} $\spm{\phantom{0}28.87}$ & \phantom{0}{\bf{403.32}} $\spm{\phantom{0}15.40}$ & \phantom{0}{\bf{403.42}} $\spm{\phantom{0}40.56}$ & \phantom{0}{\bf{449.36}} $\spm{\phantom{0}16.80}$ \\ 
    \bottomrule
\end{tabular}}
\end{table}

\begin{figure}[t]
\centering
\includegraphics[width=0.30\linewidth]{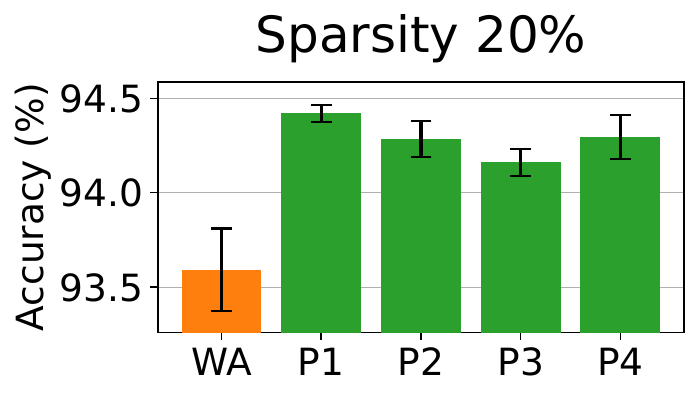}\hfill
\includegraphics[width=0.30\linewidth]{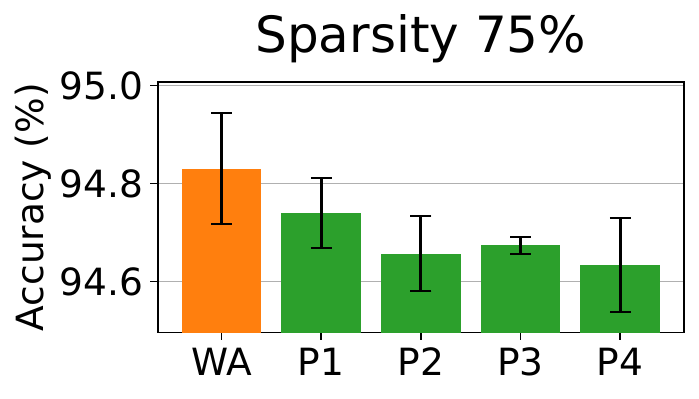}\hfill
\includegraphics[width=0.30\linewidth]{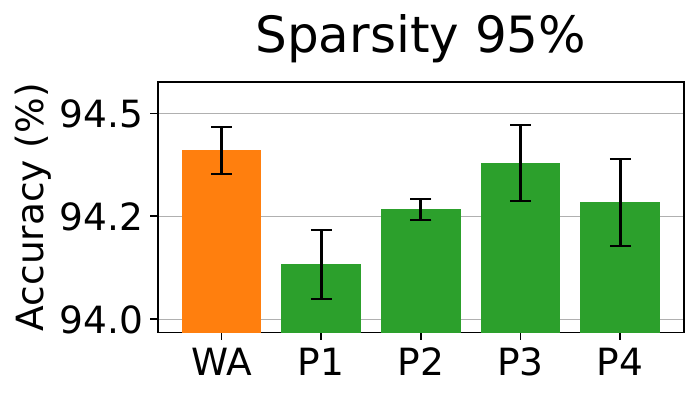}
\caption{Bar plots depicting the accuracy of individual particles involved in the averaging process of the SWAMP algorithm. While the averaged weight (denoted as WA) may not outperform individual particles (denoted as P1-P4) in the early stages of IMP (left; Sparsity 20\%), it achieves high performance at higher sparsity levels. The results are presented for WRN-28-2 on the test split of CIFAR-10. We refer readers to \cref{app:sec:additional_experiments:particle} for the same plot for CIFAR-100.}
\label{figure/particle}
\end{figure}


\begin{figure}[t]
    \centering
    \includegraphics[width=1.0\linewidth]{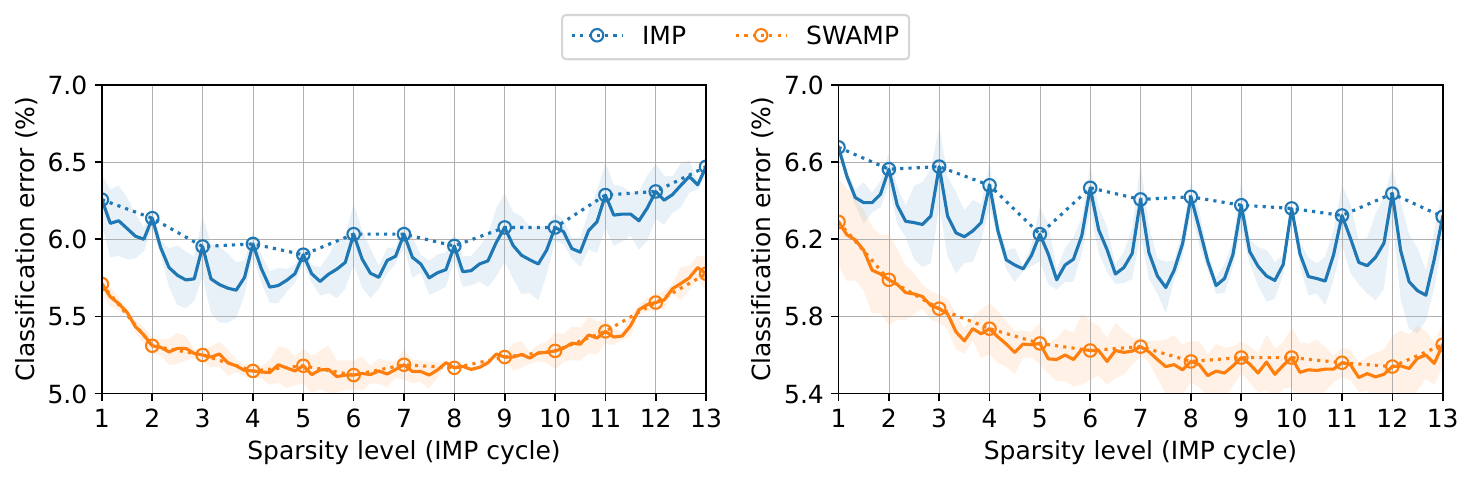}
    \caption{Linear connectivity between sparse solutions with different sparsity levels gathered from the end of IMP cycles. The results are presented for WRN-28-2 (left) and VGG-13 (right) on the test split of CIFAR-10, and we refer the reader to \cref{app:sec:additional_experiments:connectivity} for the same plot for CIFAR-100.}
    \label{figure/loss_barrier}
\end{figure}


\subsection{SWAMP: a loss landscape perspective}
\label{main:subsec:swamp_lmc}
In this section, we explore step-by-step whether the characteristics of IMP introduced in \cref{main:subsec:imp} also hold for SWAMP along with highlighting the strengths of SWAMP. To begin with, we examine the linear connectivity of SWAMP particles in a single cycle. Although \cite{frankle2020linear} empirically proves pair-wise linear connectivity, it remains uncertain whether this holds true for the convex combination of more than two particles. In \cref{figure/loss_surface}, we visualize the loss surface of IMP-trained particles along with the weight-averaged particle. We can notice that weight averaging fails at the earlier stages of IMP due to the highly non-convex nature of the landscape. However, as sparsity increases, particles tend to locate in the same wide basin which enables weight-averaging. Such a finding is in line with \cite{frankle2020linear} that demonstrated the ease of finding a low-loss curve with a smaller network compared to a larger one, i.e., a sparse network tends to be more stable. Additionally, it further demonstrates that our algorithm benefits more with sparser networks. 

\cref{figure/particle} provides additional evidence showing that the weight-averaged solution is indeed superior to its individual members, other than in the cases where the dense network is not yet stabilized. Better, the notable performance gap between individual particles promotes the need for weight-averaging. We further quantify the flatness of the local minima through the trace of Hessian employing the Power Iteration algorithm~\citep{yao2020pyhessian}. Higher Hessian trace value implies that the converged local minimum exhibits a high curvature. The results in \cref{table/main_hessian_trace} validate that SWAMP locates a flatter sparse network compared to IMP, only except for earlier cycles. 

Finally, we check whether consecutive solutions from SWAMP cycles are linearly connected -- a key to the success of IMP pointed out by \cite{paul2023unmasking} -- which indeed turns out to be true according to \cref{figure/loss_barrier}. Not only is this true, but our method also exhibits minimal variance and maintains a highly stable trajectory throughout the pruning process, suggesting that SWAMP finds a flat and well-connected basin. To this end, we provide empirical evidence that our method effectively identifies flat minima while retaining the desirable properties of IMP, resulting in a single highly sparse network that outperforms IMP.

\section{Experiments}
\label{main:sec:experiments}


\subsection{Main results: image classification tasks}
\label{main:subsec:main_results}

\begin{table}[t]
\caption{Classification accuracy on residual networks. SWAMP outperforms all the baselines across varying sparsities. Reported values are averaged over three random seeds, and the best and second-best results are boldfaced and underlined, respectively.}
\label{table/main_cifar10}
\setlength{\tabcolsep}{0.9em}
\centering
\resizebox{0.90\linewidth}{!}{\begin{tabular}{clcccc}
    \toprule
    &        & \multicolumn{4}{c}{Sparsity}\\
    \cmidrule(lr){3-6}
    & Method & 50\% & 75\% & 90\% & 95\% \\
    \midrule
    \multirow{9}{*}{\Centerstack{CIFAR-10\\(WRN-28-2)}}
        & SNIP              & 92.63$\spm{0.56}$ & 92.13$\spm{0.21}$ & 91.02$\spm{0.81}$ & 88.97$\spm{0.35}$ \\
        & SynFlow           & 92.99$\spm{0.46}$ & 92.98$\spm{0.08}$ & 90.98$\spm{0.30}$ & 89.53$\spm{0.20}$ \\
        & GraSP             & 92.34$\spm{0.26}$ & 90.74$\spm{0.27}$ & 89.70$\spm{0.60}$ & 88.78$\spm{0.22}$ \\
        & RigL              & 93.59$\spm{0.15}$ & 93.09$\spm{0.13}$ & 91.81$\spm{0.18}$ & 90.73$\spm{0.16}$ \\
        & DST               & 94.04$\spm{0.37}$ & 93.62$\spm{0.12}$ & 92.59$\spm{0.40}$ & 92.10$\spm{0.34}$ \\
        & IMP               & 93.97$\spm{0.16}$ & 94.02$\spm{0.23}$ & 93.90$\spm{0.15}$ & \underline{93.58}$\spm{0.09}$ \\
        & IMP + SAM         & 94.06$\spm{0.30}$ & 94.20$\spm{0.54}$ & 93.89$\spm{0.53}$ & 92.03$\spm{1.93}$ \\
        & Lottery Pools     & \underline{94.39}$\spm{0.16}$ & \underline{94.28}$\spm{0.14}$ & \underline{94.16}$\spm{0.11}$ & 93.43$\spm{0.23}$ \\
        \cmidrule(lr){2-6}
        & {\bf SWAMP (ours)} & {\bf 94.74}$\spm{0.04}$ & {\bf 94.88}$\spm{0.09}$ & {\bf 94.73}$\spm{0.10}$ & {\bf 94.23}$\spm{0.11}$ \\
    \midrule   
    \multirow{9}{*}{\Centerstack{CIFAR-100\\(WRN-32-4)}}
        & SNIP              & 71.98$\spm{0.37}$ & 72.16$\spm{0.77}$ & 69.82$\spm{1.14}$ & 68.10$\spm{0.28}$ \\
        & SynFlow           & 75.16$\spm{0.54}$ & 74.22$\spm{0.60}$ & 74.60$\spm{0.54}$ & 74.17$\spm{0.77}$ \\
        & GraSP             & 70.75$\spm{1.78}$ & 68.17$\spm{0.48}$ & 66.96$\spm{0.05}$ & 65.80$\spm{0.89}$ \\
        & RigL              & 75.05$\spm{0.48}$ & 73.37$\spm{0.11}$ & 71.23$\spm{0.76}$ & 70.23$\spm{0.48}$ \\
        & DST               & 74.78$\spm{1.23}$ & 74.02$\spm{1.73}$ & 72.79$\spm{1.16}$ & 71.41$\spm{0.43}$ \\
        & IMP               & 75.40$\spm{0.23}$ & 75.72$\spm{0.41}$ & 75.24$\spm{0.25}$ & 74.60$\spm{0.37}$ \\
        & IMP + SAM         & 75.63$\spm{0.70}$ & \underline{76.19}$\spm{0.81}$ & \underline{75.85}$\spm{0.78}$ & 75.07$\spm{0.65}$ \\
        & Lottery Pools     & \underline{76.31}$\spm{0.51}$ & 76.17$\spm{1.03}$ & 75.84$\spm{0.67}$ & \underline{75.14}$\spm{0.49}$ \\
        \cmidrule(lr){2-6}
        & {\bf SWAMP (ours)} & {\bf 77.29}$\spm{0.53}$ & {\bf 77.35}$\spm{0.39}$ & {\bf 77.14}$\spm{0.33}$ & {\bf 76.48}$\spm{0.73}$ \\

    \bottomrule
\end{tabular}}
\end{table}

\begin{table}[t]
\centering
\caption{Classification accuracy with ResNet-50, which is trained on ImageNet-Train, validated on IN-Valid, and tested on ImageNet-V2, ImageNet-Rendition (denoted by IN-R), and ImageNet-Sketch (denoted by IN-S). Reported values are averaged over three trials.}
\label{table/rebuttal_imagenet}
\setlength{\tabcolsep}{0.5em}
\resizebox{0.85\textwidth}{!}{
\begin{tabular}{clllllll}
    \toprule
    & & \multicolumn{6}{c}{Sparsity} \\
    \cmidrule(lr){3-8}
    & Method & 0\% & 45.9\% & 68.8\% & 80.3\% & 86.0\% & 88.9 \% \\
    \midrule
    \multirow{2}{*}{\Centerstack{IN-Val}}
    & IMP        & 76.25$\spm{0.04}$ & 76.40$\spm{0.10}$ & 76.13$\spm{0.09}$ & 75.54$\spm{0.07}$ & 74.22$\spm{0.09}$ & 71.66$\spm{0.09}$ \\
    & \bf{SWAMP} & -                 & 76.56$\spm{0.08}$ & 76.51$\spm{0.03}$ & 75.69$\spm{0.20}$ & 74.25$\spm{0.03}$ & 71.81$\spm{0.19}$ \\
    \midrule
    \multirow{2}{*}{\Centerstack{IN-V2}}
    & IMP        & 64.18$\spm{0.10}$ & 64.11$\spm{0.12}$ & 63.78$\spm{0.03}$ & 62.77$\spm{0.19}$ & 61.20$\spm{0.19}$ & 59.00$\spm{0.32}$ \\
    & \bf{SWAMP} & -                 & 64.34$\spm{0.38}$ & 64.06$\spm{0.40}$ & 63.43$\spm{0.24}$ & 61.82$\spm{0.15}$ & 59.44$\spm{0.07}$ \\
    \midrule
    \multirow{2}{*}{\Centerstack{IN-R}}
    & IMP        & 35.38$\spm{0.27}$ & 35.04$\spm{0.25}$ & 34.71$\spm{0.19}$ & 34.05$\spm{0.26}$ & 32.85$\spm{0.29}$ & 30.92$\spm{0.10}$ \\
    & \bf{SWAMP} & -                 & 37.12$\spm{0.11}$ & 36.61$\spm{0.16}$ & 35.61$\spm{0.22}$ & 34.14$\spm{0.31}$ & 32.21$\spm{0.73}$ \\
    \midrule
    \multirow{2}{*}{\Centerstack{IN-S}}
    & IMP        & 23.90$\spm{0.02}$ & 23.86$\spm{0.23}$ & 23.74$\spm{0.32}$ & 22.88$\spm{0.26}$ & 21.60$\spm{0.24}$ & 19.32$\spm{0.46}$ \\
    & \bf{SWAMP} & -                 & 25.29$\spm{0.18}$ & 24.95$\spm{0.17}$ & 24.05$\spm{0.33}$ & 22.43$\spm{0.02}$ & 20.08$\spm{0.42}$ \\
    \bottomrule
\end{tabular}}
\end{table}

\textbf{Baseline approaches.}
In addition to IMP with weight rewinding~\citep{frankle2020linear}, our method is compared to a list of pruning techniques. This includes one-shot pruning methods (SNIP~\citep{lee2018snip}; GraSP~\citep{wang2020picking}; SynFlow~\citep{tanaka2020pruning}), dense-to-sparse training with dynamic masks (DST~\citep{liu2020dynamic}; RigL~\citep{evci2020rigging}), SAM-optimized IMP~\citep{na2022train}, and Lottery Pools~\citep{yin2023lottery}. 

\textbf{Experimental setup.}
Our method is evaluated on diverse image classification benchmarks, which include CIFAR-10, CIFAR-100, Tiny-ImageNet and ImageNet datasets. Throughout the experiments, we use residual networks~\citep{he2016deep} and VGG networks~\citep{Karen2015vgg} as a basis: WRN-28-2 and VGG-13 for CIFAR-10; WRN-32-4 and VGG-16 for CIFAR-100; R18 for Tiny-ImageNet; and R50 for ImageNet. Unless specified, we set the number of SWAMP particles $N=4$ and pruning ratio $\alpha=0.2$. Refer to \cref{app:sec:details} for further experimental details. 

\cref{table/main_cifar10} presents the performance of SWAMP together with other baseline methods on the CIFAR-10 and CIFAR-100 datasets, respectively. Compared to other baseline methods, SWAMP consistently achieves the highest classification accuracy across all sparsity levels and models when evaluated on the CIFAR-10 and CIFAR-100 test sets; we defer results on VGG networks and Tiny-ImageNet to \cref{app:sec:additional_experiments}. To see the uncertainty quantification aspects, in \cref{app:sec:additional_experiments:sparsity}, we report negative log-likelihoods (NLLs) as well. Again, SWAMP achieves the best NLL in all settings. Furthermore, \cref{table/rebuttal_imagenet} highlights that our method consistently outperforms IMP on ImageNet, a large-scale dataset that is known to be hard to prune. 



\subsection{Ablation studies}
\label{main:subsec:ablation}

\begin{table}[t]
\caption{Ablation study to validate SWAMP's efficacy in two aspects: (i) mask generation and (ii) sparse training. Except for earlier iterations, i.e., low sparsity regime, our method excels in both areas relative to vanilla SGD optimization. Reported values are classification accuracy averaged over three random seeds, and the best results are boldfaced. Refer to \cref{app:sec:additional_experiments:ablation} for VGG networks.}
\label{table:mask_and_training}
\setlength{\tabcolsep}{0.9em}
\centering
\resizebox{0.9\linewidth}{!}{\begin{tabular}{cllcccc}
    \toprule
     & & & \multicolumn{4}{c}{Sparsity} \\
     \cmidrule(lr){4-7}
     & Mask & Training & 50\% & 75\% & 90\% & 95\%  \\
     \midrule
    \multirow{4}{*}{\Centerstack{CIFAR-10\\(WRN-28-2)}}
        & SGD   & SGD   & 93.97$\spm{0.16}$ & 94.02$\spm{0.23}$ & 93.90$\spm{0.15}$ & 93.58$\spm{0.09}$\\ 
        & SGD   & SWAMP & {\bf 94.85}$\spm{0.05}$ & {\bf 94.91}$\spm{0.09}$ & 94.48$\spm{0.06}$ & 93.99$\spm{0.31}$\\ 
        \cmidrule(lr){2-7}
        & SWAMP & SGD & 94.15$\spm{0.15}$ & 94.37$\spm{0.18}$ & 94.26$\spm{0.03}$ & 93.83$\spm{0.16}$\\ 
        & SWAMP & SWAMP & 94.74$\spm{0.04}$ & 94.88$\spm{0.09}$ & {\bf 94.73}$\spm{0.10}$ & {\bf 94.23}$\spm{0.11}$ \\ 
    \bottomrule
\end{tabular}}
\end{table}

\begin{table}[t]
\centering  
\caption{Ablation study on the impact of the two main components of SWAMP; averaging multiple particles (denoted by Multi) and averaging learning trajectory (denoted by SWA). Our findings indicate that the best performance is achieved when both techniques are employed. Reported classification accuracies are averaged over three random seeds, and the best and second-best results are boldfaced and underlined, respectively. Refer to \cref{app:sec:additional_experiments:ablation} for VGG networks.}
\label{table:main_sgd_swa}
\setlength{\tabcolsep}{0.9em}
\resizebox{0.85\linewidth}{!}{\begin{tabular}{ccccccc}
    \toprule
    & & & \multicolumn{4}{c}{Sparsity} \\
    \cmidrule(lr){4-7}
    & Multi & SWA & 50\% & 75\% & 90\% & 95\%  \\
    \midrule
    \multirow{4}{*}{\Centerstack{CIFAR-10\\(WRN-28-2)}}
        & \cmark & \cmark 
        & {\bf 94.74}$\spm{0.04}$ & 
        {\bf 94.88}$\spm{0.09}$ & 
        {\bf 94.73}$\spm{0.10}$ & 
        {\bf 94.23}$\spm{0.11}$ \\ 
        & \cmark & \xmark 
        & 94.43$\spm{0.10}$ & 
        94.44$\spm{0.19}$ & 
        \underline{94.37}$\spm{0.12}$ & 
        93.73$\spm{0.32}$ \\
        & \xmark & \cmark 
        & \underline{94.62}$\spm{0.06}$ & 
        \underline{94.67}$\spm{0.06}$ & 
        94.35$\spm{0.06}$ & 
        \underline{93.97}$\spm{0.10}$ \\ 
        & \xmark & \xmark 
        & 93.97$\spm{0.16}$ & 
        94.02$\spm{0.23}$ & 
        93.90$\spm{0.15}$ & 
        93.58$\spm{0.09}$\\ 
    \bottomrule
\end{tabular}}
\end{table}

\begin{table}[ht]
\caption{Ablation study on the number of SWAMP particles. The finding indicates that the performance improves with an increase in the number of particles. Reported values are classification accuracy averaged over three random seeds, and the best and second-best results are boldfaced and underlined, respectively. Refer to \cref{app:sec:additional_experiments:ablation} for VGG networks.}
\label{table/main_number_of_particles}
\setlength{\tabcolsep}{1.0em}
\centering
\resizebox{0.85\linewidth}{!}{\begin{tabular}{cccccc}
    \toprule
    & & \multicolumn{4}{c}{Sparsity}\\
    \cmidrule(lr){3-6}
    & \# particles & 50\% & 75\% & 90\% & 95\%  \\
    \midrule
    \multirow{4}{*}{\Centerstack{CIFAR-10\\(WRN-28-2)}}
        & 1 & 94.62$\spm{0.06}$ & 
            94.67$\spm{0.06}$ & 
            94.35$\spm{0.06}$ & 
            93.97$\spm{0.10}$\\
        & 2 & 94.57$\spm{0.04}$ & 
            94.59$\spm{0.07}$ & 
            94.38$\spm{0.17}$ & 
            94.05$\spm{0.16}$ \\
        & 4 & \underline{94.74}$\spm{0.04}$ & 
            \underline{94.88}$\spm{0.09}$ & 
            \underline{94.73}$\spm{0.10}$ & 
            {\bf 94.23}$\spm{0.11}$ \\
        & 8 & {\bf 94.80}$\spm{0.04}$ & 
            {\bf 94.90}$\spm{0.09}$ & 
            {\bf 94.74}$\spm{0.10}$ & 
            \underline{94.21}$\spm{0.24}$ \\
    \bottomrule
\end{tabular}}
\end{table}

\textbf{Does SWAMP find a better mask?}
To validate that SWAMP indeed identifies a superior mask compared to IMP, we conduct an experiment with two different masks: (i) a mask obtained from IMP, and (ii) a mask obtained from SWAMP. At a predetermined fixed sparsity level, we initially train our model using SGD (or SWAMP) and the mask from IMP. At the same time, we train the model using SGD (or SWAMP) and the mask from SWAMP at the same sparsity level. These two processes differ only in the masks utilized, while the training approach and the fixed sparsity level remain the same. \cref{table:mask_and_training} presents clear evidence that the SWAMP mask consistently outperforms its counterpart in terms of performance, with the exception of the WRN-28-2 model at 50\% and 75\% sparsity levels when trained using SWAMP. This result shows that SWAMP generates a \textit{better} sparse mask than IMP.

\textbf{Does SWAMP offer a better sparse training?}
In \cref{table:mask_and_training}, we can also verify that SWAMP offers better sparse training compared to IMP. By comparing the results between SGD training and SWAMP training using the same mask, it becomes evident that SWAMP consistently outperforms across all masks, sparsity levels, and models. It verifies that SWAMP effectively guides the weight particles towards converging into flat local minima, resulting in improved generalization on the test split. The induced flatness of the local minima through SWAMP's weight distribution contributes to enhanced performance and robustness of the model as we discussed in \cref{main:subsec:swamp_lmc}.

\textbf{Two averaging strategies of SWAMP.}
As described in \cref{main:subsec:swamp_algorithm}, we here investigate how stochastic weight averaging and multi-particle averaging contribute to the final performance of SWAMP. In \cref{table:main_sgd_swa}, throughout all four sparsities, applying only one of the two techniques displays better performance than IMP (bottom-row) but clearly lower than SWAMP (top-row). We conclude that two ingredients complement each other, achieving optimal performance when applied together. 
Further in \cref{table/main_number_of_particles}, we conduct an empirical analysis to investigate the correlation between the number of particles and the performance of SWAMP. We provide additional ablation studies in \cref{app:sec:additional_experiments:ablation}.


\subsection{Further remarks}
\label{main:subsec:remarks}

\begin{table}[t]
    \centering
    \caption{Further comparison between (i) IMP, (ii) SWAMP, and (iii) the cost-efficient version of the SWAMP algorithm (denoted by SWAMP+) in terms of accuracy and total training FLOPs. Reported values are averaged over  three random seeds.}
    \label{table/main_later_phases}
    \setlength{\tabcolsep}{1.2em}
    \resizebox{0.85\linewidth}{!}{\begin{tabular}{clcccc}
    \toprule
    & & \multicolumn{2}{c}{Sparsity 95\%} & \multicolumn{2}{c}{Sparsity 98\%} \\
    \cmidrule(lr){3-4}\cmidrule(lr){5-6}
    & Method & Accuracy & GFLOPs & Accuracy & GFLOPs \\
    \midrule
    \multirow{3}{*}{\Centerstack{CIFAR-10\\(WRN-28-2)}} &
      IMP     & 93.58$\spm{0.09}$ & \phantom{0}1.19 & 89.05$\spm{1.39}$ & \phantom{0}1.23 \\
    & SWAMP   & 94.23$\spm{0.11}$ & \phantom{0}4.75 & 90.85$\spm{0.47}$ & \phantom{0}4.92 \\
    & SWAMP+  & 94.32$\spm{0.24}$ & \phantom{0}1.39 & 90.51$\spm{0.08}$ & \phantom{0}1.56 \\
    \midrule
    \multirow{3}{*}{\Centerstack{CIFAR-100\\(WRN-32-4)}} &
      IMP     & 74.60$\spm{0.37}$ & \phantom{0}5.25 & 70.74$\spm{0.71}$ & \phantom{0}5.38 \\
    & SWAMP   & 76.48$\spm{0.73}$ &           20.99 & 72.14$\spm{0.58}$ &           21.53 \\
    & SWAMP+  & 76.19$\spm{0.18}$ & \phantom{0}5.96 & 71.90$\spm{0.72}$ & \phantom{0}6.51 \\
    \bottomrule
    \end{tabular}}
\end{table}

\begin{table}[ht]
    \centering
    \caption{Results on CIFAR for larger pruning ratio of $\alpha \in \{0.2, 0.3, 0.4, 0.5\}$ after $13, 8, 6, 4$ IMP cycles, respectively. Reported values are averaged over  three random seeds. The numbers within parentheses indicate the difference from the IMP baseline.}
    \label{table/main_robustness}
    \resizebox{0.9\linewidth}{!}{\begin{tabular}{clllll}
    \toprule
    & Method & $\alpha=0.2$ & $\alpha=0.3$ & $\alpha=0.4$ & $\alpha=0.5$ \\
    \midrule
    \multirow{2}{*}{\Centerstack{CIFAR-10\\(WRN-28-2)}}
    & IMP   & 93.58$\spm{0.09}$ & 93.36$\spm{0.35}$ & 93.11$\spm{0.13}$ & 93.77$\spm{0.12}$ \\
    & \bf{SWAMP} & 94.23$\spm{0.11}$ \CP{0.65} & 93.83$\spm{0.04}$ \CP{0.47} & 94.41$\spm{0.02}$ \CP{1.30} & 94.68$\spm{0.04}$ \CP{0.91}\\
    \midrule
    \multirow{2}{*}{\Centerstack{CIFAR-100\\(WRN-32-4)}}
    & IMP   & 74.60$\spm{0.37}$ & 74.24$\spm{0.33}$ & 73.87$\spm{0.86}$ & 73.85$\spm{0.83}$ \\
    & \bf{SWAMP} & 76.48$\spm{0.73}$ \CP{1.88} & 77.31$\spm{0.71}$ \CP{3.07} & 76.26$\spm{0.55}$ \CP{2.39} & 76.50$\spm{0.40}$ \CP{2.65}\\
    \bottomrule
    \end{tabular}}
\end{table}

\begin{table}[!ht]
\centering
\caption{F1 score and accuracy on the development set of MRPC and RTE, respectively. SWAMP displays improved performance on language tasks as well. Reported values are averaged over three random seeds, and the numbers within parentheses indicate the difference from the IMP baseline.}
\label{table/main_roberta}
\setlength{\tabcolsep}{0.4em}
\resizebox{0.85\textwidth}{!}{
\begin{tabular}{clllll}
    \toprule
    & & \multicolumn{4}{c}{Sparsity} \\
    \cmidrule(lr){3-6}
    & Method & 0\% & 40.9\% & 52.2\% & 61.3\% \\
    \midrule
    \multirow{2}{*}{\Centerstack{MRPC\\(RoBERTa)}}
    & IMP        & 92.60$\spm{0.70}$ & 91.02$\spm{0.31}$ & 88.31$\spm{0.17}$ & 86.81$\spm{0.50}$ \\
    & \bf{SWAMP} & 92.81$\spm{0.38}$ \CP{0.21} & 91.51$\spm{0.39}$ \CP{0.49} & 89.40$\spm{0.40}$ \CP{1.09} & 88.12$\spm{0.48}$ \CP{1.31} \\
    \midrule
    \multirow{2}{*}{\Centerstack{RTE\\(RoBERTa)}}
    & IMP        & 80.13$\spm{1.02}$ & 73.25$\spm{1.52}$ & 63.30$\spm{0.37}$ & 57.46$\spm{1.39}$ \\
    & \bf{SWAMP} & 80.30$\spm{0.19}$ \CP{0.17} & 75.40$\spm{1.45}$ \CP{2.15} & 68.91$\spm{1.08}$ \CP{5.61} & 62.63$\spm{1.30}$ \CP{5.17} \\
    \bottomrule
\end{tabular}}
\end{table}

{\textbf{Parallelization strategy in distributed training.}
A notable limitation of the SWAMP algorithm is its training cost, which scales linearly with the number of particles. The training cost is typically not a major issue when working with small datasets like CIFAR-10/100, but it becomes significant when handling large datasets like ImageNet. Consequently, we suggest parallelizing the particles across machines when implemented within distributed training environments, a common practice for handling large-scale models and datasets. This strategy incurs virtually no additional costs compared to IMP, except for the extra memory required for storing the averaged parameters. Indeed, we put this strategy into practice during our ImageNet experiments, and SWAMP achieved outstanding results while incurring almost the same training expenses as IMP (cf. \cref{table/rebuttal_imagenet}).
}

{\textbf{Reducing training costs of SWAMP.}
The parallelization strategy mentioned above is exclusively applicable to distributed training setups. Consequently, we further introduce practical methods to reduce the training costs of the SWAMP algorithm, which can be used even in non-distributed training environments:
\textit{(1) Employing multiple particles only in the high-sparsity regime} mitigates the significant training costs mainly encountered in low-sparsity regimes. \cref{table/main_later_phases} demonstrates that this approach reduces training costs by a factor of 3 to 4 with minimal performance degradation. Here, we initiate training with a single-particle SWAMP for the first ten IMP cycles, achieving a sparsity level of 90\%, and then transition to using four particles afterward.
\textit{(2) Increasing the pruning ratio} decreases the number of pruning cycles necessary to achieve a certain sparsity level and thus significantly reduces total training costs. \cref{table/main_robustness} verifies that SWAMP is proficient in pruning even when using a higher pruning ratio, highlighting a distinctive advantage of SWAMP compared to IMP. These findings, combined with those in \cref{table/main_hessian_trace}, align with Lemma 3.1 in \cite{paul2023unmasking}; a smaller Hessian eigenvalue, indicating flatter minima, enhances robustness to SGD noise and making it more likely to restore matching performance.

{\textbf{Extension to language tasks and dynamic pruning.}
Up to this point, we have demonstrated that multi-particle averaging benefits IMP in iamge classification tasks. However, it is important to note that SWAMP can be easily applied to different pruning techniques across a range of tasks. To clarify, we present two distinct extensions: (1) \cref{table/main_roberta} further confirms that SWAMP outperforms IMP in language tasks, where experimental details are available in \cref{app:sec:additional_experiments:nlp}. (2) We also illustrate how SWAMP has the potential to enhance dynamic sparse training methods, along with additional results with RigL~\citep{evci2020rigging} in \cref{app:sec:additional_experiments:rigl}.
}

}





\section{Conclusion}
\label{main:sec:conclusion}

{Drawing inspiration from previous research on the relationship between iterative magnitude pruning and linear mode connectivity, we extended the single-particle scenario to incorporate multiple particles. Our initial empirical findings demonstrated that multiple models trained with different SGD noise but sharing the same matching ticket can be weight-averaged without encountering loss barriers. We further observed that the averaged particle results in flat minima with improved generalization performance. In light of these insights, we introduced SWAMP, a novel iterative global pruning technique. We also established that SWAMP preserves the linear connectivity between consecutive solutions, a critical factor contributing to the effectiveness of IMP. Extensive experiments showed that SWAMP generates superior sparse masks and effectively trains sparse networks over other baseline methods.
A theoretical analysis exploring why the convex hull of the particles in weight space forms a low-loss subspace would be a valuable direction for future research. Further, investigating the underlying principles and mathematical properties of the convex hull of the solution particles and its relationship to the low-loss subspace could provide insights into the behavior and effectiveness of the SWAMP algorithm.
}
 



\newpage
\appendix

\section*{Acknowledgement}
This work was supported by Institute of Information \& communications Technology Planning \& Evaluation (IITP) grant funded by the Korea government (MSIT) (No.2019-0-00075, Artificial Intelligence Graduate School Program (KAIST), No.2022-0-00713, Meta-learning Applicable to Real-world Problems, No.2022-0-00184, Development and Study of AI Technologies to Inexpensively Conform to Evolving Policy on Ethics), and the National Research Foundation of Korea (NRF) grant funded by the Korea government (MSIT) (NRF-2022R1A5A708390812). 

\section*{Reproducibility Statement}
Our algorithm is built on Pytorch 1.10.2~\citep{Paszke_PyTorch_An_Imperative_2019}, which is available under a BSD-style license\footnote{\href{https://github.com/pytorch/pytorch/blob/main/LICENSE}{https://github.com/pytorch/pytorch/blob/main/LICENSE}}. All experiments are conducted on NVIDIA RTX 2080 and NVIDIA RTX 3090 machines. Basic experimental setup including network and dataset choice is listed in \cref{main:subsec:main_results}. In \cref{app:sec:details}, we further provide all detailed hyperparameter settings to ensure the fair comparison between the baselines and our algorithm. We also include our code in the supplementary material. 

\section*{Ethics Statement}
The paper does not raise any ethical concerns. We only utilize publicly available datasets and python packages adhering to the appropriate licenses.

\bibliography{references}

\begin{thebibliography}{54}
\providecommand{\natexlab}[1]{#1}
\providecommand{\url}[1]{\texttt{#1}}
\expandafter\ifx\csname urlstyle\endcsname\relax
  \providecommand{\doi}[1]{doi: #1}\else
  \providecommand{\doi}{doi: \begingroup \urlstyle{rm}\Url}\fi

\bibitem[Ashukha et~al.(2020)Ashukha, Lyzhov, Molchanov, and Vetrov]{ashukha2020pitfalls}
Arsenii Ashukha, Alexander Lyzhov, Dmitry Molchanov, and Dmitry Vetrov.
\newblock Pitfalls of in-domain uncertainty estimation and ensembling in deep learning.
\newblock In \emph{International Conference on Learning Representations (ICLR)}, 2020.

\bibitem[Bar~Haim et~al.(2006)Bar~Haim, Dagan, Dolan, Ferro, Giampiccolo, Magnini, and Szpektor]{bar2006second}
Roy Bar~Haim, Ido Dagan, Bill Dolan, Lisa Ferro, Danilo Giampiccolo, Bernardo Magnini, and Idan Szpektor.
\newblock The second {PASCAL} recognising textual entailment challenge, 2006.

\bibitem[Bentivogli et~al.(2009)Bentivogli, Dagan, Dang, Giampiccolo, and Magnini]{bentivogli2009fifth}
Luisa Bentivogli, Ido Dagan, Hoa~Trang Dang, Danilo Giampiccolo, and Bernardo Magnini.
\newblock The fifth {PASCAL} recognizing textual entailment challenge, 2009.

\bibitem[Benton et~al.(2021)Benton, Maddox, Lotfi, and Wilson]{benton2021loss}
G.~W. Benton, W.~J. Maddox, S.~Lotfi, and A.~G. Wilson.
\newblock Loss surface simplexes for mode connecting volumes ans fast ensembling.
\newblock In \emph{Proceedings of the 38th International Conference on Machine Learning (ICML)}, 2021.

\bibitem[Chaudhari et~al.(2017)Chaudhari, Choromanska, Soatto, LeCun, Baldassi, Borgs, Chayes, Sagun, and Zecchina]{chaudhari2017entropysgd}
Pratik Chaudhari, Anna Choromanska, Stefano Soatto, Yann LeCun, Carlo Baldassi, Christian Borgs, Jennifer Chayes, Levent Sagun, and Riccardo Zecchina.
\newblock Entropy-{SGD}: Biasing gradient descent into wide valleys.
\newblock In \emph{International Conference on Learning Representations (ICLR)}, 2017.

\bibitem[Chen et~al.(2021)Chen, Sui, Chen, Zhang, and Wang]{chen2021unified}
Tianlong Chen, Yongduo Sui, Xuxi Chen, Aston Zhang, and Zhangyang Wang.
\newblock A unified lottery ticket hypothesis for graph neural networks.
\newblock In \emph{Proceedings of the 38th International Conference on Machine Learning (ICML)}, 2021.

\bibitem[Dagan et~al.(2006)Dagan, Glickman, and Magnini]{dagan2006pascal}
Ido Dagan, Oren Glickman, and Bernardo Magnini.
\newblock The {PASCAL} recognising textual entailment challenge.
\newblock In \emph{Machine learning challenges. evaluating predictive uncertainty, visual object classification, and recognising tectual entailment}, pp.\  177--190. Springer, 2006.

\bibitem[Dolan \& Brockett(2005)Dolan and Brockett]{dolan2005automatically}
William~B Dolan and Chris Brockett.
\newblock Automatically constructing a corpus of sentential paraphrases.
\newblock In \emph{Proceedings of the International Workshop on Paraphrasing}, 2005.

\bibitem[Dong et~al.(2017)Dong, Chen, and Pan]{dong2017learning}
Xin Dong, Shangyu Chen, and Sinno Pan.
\newblock Learning to prune deep neural networks via layer-wise optimal brain surgeon.
\newblock In \emph{Advances in Neural Information Processing Systems (NIPS)}, 2017.

\bibitem[Draxler et~al.(2018)Draxler, Veschgini, Salmhofer, and Hamprecht]{draxler2018essentially}
Felix Draxler, Kambis Veschgini, Manfred Salmhofer, and Fred Hamprecht.
\newblock Essentially no barriers in neural network energy landscape.
\newblock In \emph{Proceedings of the 35th International Conference on Machine Learning (ICML)}, 2018.

\bibitem[Evci et~al.(2020)Evci, Gale, Menick, Castro, and Elsen]{evci2020rigging}
Utku Evci, Trevor Gale, Jacob Menick, Pablo~Samuel Castro, and Erich Elsen.
\newblock Rigging the lottery: Making all tickets winners.
\newblock In \emph{Proceedings of the 37th International Conference on Machine Learning (ICML)}, 2020.

\bibitem[Evci et~al.(2022)Evci, Ioannou, Keskin, and Dauphin]{evci2022gradient}
Utku Evci, Yani Ioannou, Cem Keskin, and Yann Dauphin.
\newblock Gradient flow in sparse neural networks and how lottery tickets win.
\newblock \emph{Proceedings of the AAAI Conference on Artificial Intelligence}, 36\penalty0 (6), Jun. 2022.

\bibitem[Foret et~al.(2021)Foret, Kleiner, Mobahi, and Neyshabur]{foret2021sharpnessaware}
Pierre Foret, Ariel Kleiner, Hossein Mobahi, and Behnam Neyshabur.
\newblock Sharpness-aware minimization for efficiently improving generalization.
\newblock In \emph{International Conference on Learning Representations (ICLR)}, 2021.

\bibitem[Fort \& Jastrzebski(2019)Fort and Jastrzebski]{fort2019large}
Stanislav Fort and Stanislaw Jastrzebski.
\newblock Large scale structure of neural network loss landscapes.
\newblock In \emph{Advances in Neural Information Processing Systems (NeurIPS)}, 2019.

\bibitem[Fort et~al.(2020)Fort, Dziugaite, Paul, Kharaghani, Roy, and Ganguli]{fort2020deep}
Stanislav Fort, Gintare~Karolina Dziugaite, Mansheej Paul, Sepideh Kharaghani, Daniel~M Roy, and Surya Ganguli.
\newblock Deep learning versus kernel learning: an empirical study of loss landscape geometry and the time evolution of the neural tangent kernel.
\newblock In \emph{Advances in Neural Information Processing Systems (NeurIPS)}, 2020.

\bibitem[Frankle \& Carbin(2019)Frankle and Carbin]{frankle2019lottery}
Jonathan Frankle and Michael Carbin.
\newblock The lottery ticket hypothesis: Finding sparse, trainable neural networks.
\newblock In \emph{International Conference on Learning Representations (ICLR)}, 2019.

\bibitem[Frankle et~al.(2020)Frankle, Dziugaite, Roy, and Carbin]{frankle2020linear}
Jonathan Frankle, Gintare~Karolina Dziugaite, Daniel Roy, and Michael Carbin.
\newblock Linear mode connectivity and the lottery ticket hypothesis.
\newblock In \emph{Proceedings of the 37th International Conference on Machine Learning (ICML)}, 2020.

\bibitem[Frankle et~al.(2021)Frankle, Dziugaite, Roy, and Carbin]{frankle2020pruning}
Jonathan Frankle, Gintare~Karolina Dziugaite, Daniel Roy, and Michael Carbin.
\newblock Pruning neural networks at initialization: Why are we missing the mark?
\newblock In \emph{International Conference on Learning Representations (ICLR)}, 2021.

\bibitem[Garipov et~al.(2018)Garipov, Izmailov, Podoprikhin, Vetrov, and Wilson]{garipov2018loss}
Timur Garipov, Pavel Izmailov, Dmitrii Podoprikhin, Dmitry~P Vetrov, and Andrew~G Wilson.
\newblock Loss surfaces, mode connectivity, and fast ensembling of {DNNs}.
\newblock In \emph{Advances in Neural Information Processing Systems (NeurIPS)}, 2018.

\bibitem[Giampiccolo et~al.(2007)Giampiccolo, Magnini, Dagan, and Dolan]{giampiccolo2007third}
Danilo Giampiccolo, Bernardo Magnini, Ido Dagan, and Bill Dolan.
\newblock The third {PASCAL} recognizing textual entailment challenge.
\newblock In \emph{Proceedings of the ACL-PASCAL workshop on textual entailment and paraphrasing}, pp.\  1--9. Association for Computational Linguistics, 2007.

\bibitem[Girish et~al.(2021)Girish, Maiya, Gupta, Chen, Davis, and Shrivastava]{girish2021lottery}
Sharath Girish, Shishira~R Maiya, Kamal Gupta, Hao Chen, Larry~S Davis, and Abhinav Shrivastava.
\newblock The lottery ticket hypothesis for object recognition.
\newblock In \emph{Proceedings of the IEEE/CVF Conference on Computer Vision and Pattern Recognition (CVPR)}, 2021.

\bibitem[Guo et~al.(2017)Guo, Pleiss, Sun, and Weinberger]{guo2017calibration}
Chuan Guo, Geoff Pleiss, Yu~Sun, and Kilian~Q Weinberger.
\newblock On calibration of modern neural networks.
\newblock In \emph{Proceedings of the 34th International Conference on Machine Learning (ICML)}, 2017.

\bibitem[Han et~al.(2015)Han, Pool, Tran, and Dally]{han2015learning}
S.~Han, J.~Pool, J.~Tran, and W.~Dally.
\newblock Learning both weights and connections for efficient neural network.
\newblock In \emph{Advances in Neural Information Processing Systems (NIPS)}, 2015.

\bibitem[Hassibi \& Stork(1992)Hassibi and Stork]{hassibi1992second}
Babak Hassibi and David Stork.
\newblock Second order derivatives for network pruning: Optimal brain surgeon.
\newblock In \emph{Advances in Neural Information Processing Systems (NIPS)}, 1992.

\bibitem[He et~al.(2016)He, Zhang, Ren, and Sun]{he2016deep}
Kaiming He, Xiangyu Zhang, Shaoqing Ren, and Jian Sun.
\newblock Deep residual learning for image recognition.
\newblock In \emph{Proceedings of the IEEE Conference on Computer Vision and Pattern Recognition (CVPR)}, 2016.

\bibitem[Izmailov et~al.(2018)Izmailov, Podoprikhin, Garipov, Vetrov, and Wilson]{izmailov2018averaging}
P.~Izmailov, D.~Podoprikhin, T.~Garipov, D.~Vetrov, and A.~G. Wilson.
\newblock Averaging weights leads to wider optima and better generalization.
\newblock In \emph{Proceedings of the 34th Conference on Uncertainty in Artificial Intelligence (UAI)}, 2018.

\bibitem[Izmailov et~al.(2020)Izmailov, Maddox, Kirichenko, Garipov, Vetrov, and Wilson]{izmailov2020subspace}
Pavel Izmailov, Wesley~J Maddox, Polina Kirichenko, Timur Garipov, Dmitry Vetrov, and Andrew~Gordon Wilson.
\newblock Subspace inference for bayesian deep learning.
\newblock In \emph{Proceedings of the 36th Conference on Uncertainty in Artificial Intelligence (UAI)}, 2020.

\bibitem[Janowsky(1989)]{janowsky1989pruning}
Steven~A Janowsky.
\newblock Pruning versus clipping in neural networks.
\newblock \emph{Physical Review A}, 39\penalty0 (12):\penalty0 6600, 1989.

\bibitem[Kalibhat et~al.(2021)Kalibhat, Balaji, and Feizi]{kalibhat2021winning}
Neha~Mukund Kalibhat, Yogesh Balaji, and Soheil Feizi.
\newblock Winning lottery tickets in deep generative models.
\newblock \emph{Proceedings of the AAAI Conference on Artificial Intelligence}, 35\penalty0 (9), May 2021.

\bibitem[Kingma \& Ba(2015)Kingma and Ba]{kingma2015adam}
Diederick~P Kingma and Jimmy Ba.
\newblock Adam: A method for stochastic optimization.
\newblock In \emph{International Conference on Learning Representations (ICLR)}, 2015.

\bibitem[Lakshminarayanan et~al.(2017)Lakshminarayanan, Pritzel, and Blundell]{lakshminarayanan2017simple}
Balaji Lakshminarayanan, Alexander Pritzel, and Charles Blundell.
\newblock Simple and scalable predictive uncertainty estimation using deep ensembles.
\newblock In \emph{Advances in Neural Information Processing Systems (NIPS)}, 2017.

\bibitem[Larsen et~al.(2022)Larsen, Fort, Becker, and Ganguli]{larsen2021many}
Brett~W Larsen, Stanislav Fort, Nic Becker, and Surya Ganguli.
\newblock How many degrees of freedom do we need to train deep networks: a loss landscape perspective.
\newblock In \emph{International Conference on Learning Representations (ICLR)}, 2022.

\bibitem[LeCun et~al.(1989)LeCun, Denker, and Solla]{lecun1989optimal}
Yann LeCun, John Denker, and Sara Solla.
\newblock Optimal brain damage.
\newblock In \emph{Advances in Neural Information Processing Systems (NIPS)}, 1989.

\bibitem[Lee et~al.(2019)Lee, Ajanthan, and Torr]{lee2018snip}
Namhoon Lee, Thalaiyasingam Ajanthan, and Philip~HS Torr.
\newblock Snip: Single-shot network pruning based on connection sensitivity.
\newblock In \emph{International Conference on Learning Representations (ICLR)}, 2019.

\bibitem[Liu et~al.(2020)Liu, XU, SHI, Cheung, and So]{liu2020dynamic}
Junjie Liu, Zhe XU, Runbin SHI, Ray C.~C. Cheung, and Hayden~K.H. So.
\newblock Dynamic sparse training: Find efficient sparse network from scratch with trainable masked layers.
\newblock In \emph{International Conference on Learning Representations (ICLR)}, 2020.

\bibitem[Liu et~al.(2019)Liu, Ott, Goyal, Du, Joshi, Chen, Levy, Lewis, Zettlemoyer, and Stoyanov]{liu2019roberta}
Yinhan Liu, Myle Ott, Naman Goyal, Jingfei Du, Mandar Joshi, Danqi Chen, Omer Levy, Mike Lewis, Luke Zettlemoyer, and Veselin Stoyanov.
\newblock Roberta: A robustly optimized bert pretraining approach.
\newblock \emph{arXiv preprint arXiv:1907.11692}, 2019.

\bibitem[Ma et~al.(2021)Ma, Yuan, Shen, Chen, Chen, Chen, Liu, Qin, Liu, Wang, et~al.]{ma2021sanity}
Xiaolong Ma, Geng Yuan, Xuan Shen, Tianlong Chen, Xuxi Chen, Xiaohan Chen, Ning Liu, Minghai Qin, Sijia Liu, Zhangyang Wang, et~al.
\newblock Sanity checks for lottery tickets: Does your winning ticket really win the jackpot?
\newblock In \emph{Advances in Neural Information Processing Systems (NeurIPS)}, 2021.

\bibitem[Na et~al.(2022)Na, Mehta, and Strubell]{na2022train}
Clara Na, Sanket~Vaibhav Mehta, and Emma Strubell.
\newblock Train flat, then compress: Sharpness-aware minimization learns more compressible models.
\newblock In \emph{Findings of the Association for Computational Linguistics: EMNLP 2022}, 2022.

\bibitem[Park et~al.(2020)Park, Lee, Mo, and Shin]{park2020lookahead}
Sejun Park, Jaeho Lee, Sangwoo Mo, and Jinwoo Shin.
\newblock Lookahead: A far-sighted alternative of magnitude-based pruning.
\newblock In \emph{International Conference on Learning Representations (ICLR)}, 2020.

\bibitem[Paszke et~al.(2019)Paszke, Gross, Massa, Lerer, Bradbury, Chanan, Killeen, Lin, Gimelshein, Antiga, Desmaison, Kopf, Yang, DeVito, Raison, Tejani, Chilamkurthy, Steiner, Fang, Bai, and Chintala]{Paszke_PyTorch_An_Imperative_2019}
Adam Paszke, Sam Gross, Francisco Massa, Adam Lerer, James Bradbury, Gregory Chanan, Trevor Killeen, Zeming Lin, Natalia Gimelshein, Luca Antiga, Alban Desmaison, Andreas Kopf, Edward Yang, Zachary DeVito, Martin Raison, Alykhan Tejani, Sasank Chilamkurthy, Benoit Steiner, Lu~Fang, Junjie Bai, and Soumith Chintala.
\newblock {PyTorch: An Imperative Style, High-Performance Deep Learning Library}.
\newblock In \emph{Advances in Neural Information Processing Systems (NeurIPS)}, 2019.

\bibitem[Paul et~al.(2023)Paul, Chen, Larsen, Frankle, Ganguli, and Dziugaite]{paul2023unmasking}
Mansheej Paul, Feng Chen, Brett~W. Larsen, Jonathan Frankle, Surya Ganguli, and Gintare~Karolina Dziugaite.
\newblock Unmasking the lottery ticket hypothesis: What's encoded in a winning ticket's mask?
\newblock In \emph{International Conference on Learning Representations (ICLR)}, 2023.

\bibitem[Renda et~al.(2020)Renda, Frankle, and Carbin]{renda2020comparing}
Alex Renda, Jonathan Frankle, and Michael Carbin.
\newblock Comparing rewinding and fine-tuning in neural network pruning.
\newblock In \emph{International Conference on Learning Representations (ICLR)}, 2020.

\bibitem[Robbins \& Monro(1951)Robbins and Monro]{robbins1951stochastic}
Herbert Robbins and Sutton Monro.
\newblock A stochastic approximation method.
\newblock \emph{The annals of mathematical statistics}, pp.\  400--407, 1951.

\bibitem[Simonyan \& Zisserman(2015)Simonyan and Zisserman]{Karen2015vgg}
Karen Simonyan and Andrew Zisserman.
\newblock Very deep convolutional networks for large-scale image recognition.
\newblock In \emph{International Conference on Learning Representations (ICLR)}, 2015.

\bibitem[Su et~al.(2020)Su, Chen, Cai, Wu, Gao, Wang, and Lee]{su2020sanity}
Jingtong Su, Yihang Chen, Tianle Cai, Tianhao Wu, Ruiqi Gao, Liwei Wang, and Jason~D Lee.
\newblock Sanity-checking pruning methods: Random tickets can win the jackpot.
\newblock In \emph{Advances in Neural Information Processing Systems (NeurIPS)}, 2020.

\bibitem[Tanaka et~al.(2020)Tanaka, Kunin, Yamins, and Ganguli]{tanaka2020pruning}
Hidenori Tanaka, Daniel Kunin, Daniel~L Yamins, and Surya Ganguli.
\newblock Pruning neural networks without any data by iteratively conserving synaptic flow.
\newblock In \emph{Advances in Neural Information Processing Systems (NeurIPS)}, 2020.

\bibitem[Wang et~al.(2019)Wang, Singh, Michael, Hill, Levy, and Bowman]{wang2018glue}
Alex Wang, Amanpreet Singh, Julian Michael, Felix Hill, Omer Levy, and Samuel~R Bowman.
\newblock Glue: A multi-task benchmark and analysis platform for natural language understanding.
\newblock In \emph{International Conference on Learning Representations (ICLR)}, 2019.

\bibitem[Wang et~al.(2020)Wang, Zhang, and Grosse]{wang2020picking}
Chaoqi Wang, Guodong Zhang, and Roger Grosse.
\newblock Picking winning tickets before training by preserving gradient flow.
\newblock In \emph{International Conference on Learning Representations (ICLR)}, 2020.

\bibitem[Wortsman et~al.(2021)Wortsman, Horton, Guestrin, Farhadi, and Rastegari]{wortsman2021learning}
Mitchell Wortsman, Maxwell~C Horton, Carlos Guestrin, Ali Farhadi, and Mohammad Rastegari.
\newblock Learning neural network subspaces.
\newblock In \emph{Proceedings of the 38th International Conference on Machine Learning (ICML)}, 2021.

\bibitem[Wortsman et~al.(2022)Wortsman, Ilharco, Gadre, Roelofs, Gontijo-Lopes, Morcos, Namkoong, Farhadi, Carmon, Kornblith, et~al.]{wortsman2022model}
Mitchell Wortsman, Gabriel Ilharco, Samir~Ya Gadre, Rebecca Roelofs, Raphael Gontijo-Lopes, Ari~S Morcos, Hongseok Namkoong, Ali Farhadi, Yair Carmon, Simon Kornblith, et~al.
\newblock Model soups: averaging weights of multiple fine-tuned models improves accuracy without increasing inference time.
\newblock In \emph{Proceedings of the 39th International Conference on Machine Learning (ICML)}, 2022.

\bibitem[Yao et~al.(2020)Yao, Gholami, Keutzer, and Mahoney]{yao2020pyhessian}
Zhewei Yao, Amir Gholami, Kurt Keutzer, and Michael~W Mahoney.
\newblock Pyhessian: Neural networks through the lens of the hessian.
\newblock In \emph{2020 IEEE international conference on big data (Big data)}, 2020.

\bibitem[Yin et~al.(2023)Yin, Liu, Fang, Huang, Menkovski, and Pechenizkiy]{yin2023lottery}
Lu~Yin, Shiwei Liu, Meng Fang, Tianjin Huang, Vlado Menkovski, and Mykola Pechenizkiy.
\newblock Lottery pools: Winning more by interpolating tickets without increasing training or inference cost.
\newblock \emph{Proceedings of the AAAI Conference on Artificial Intelligence}, 37\penalty0 (9), Jun. 2023.

\bibitem[You et~al.(2020)You, Li, Xu, Fu, Wang, Chen, Baraniuk, Wang, and Lin]{you2019drawing}
Haoran You, Chaojian Li, Pengfei Xu, Yonggan Fu, Yue Wang, Xiaohan Chen, Richard~G Baraniuk, Zhangyang Wang, and Yingyan Lin.
\newblock Drawing early-bird tickets: Towards more efficient training of deep networks.
\newblock In \emph{International Conference on Learning Representations (ICLR)}, 2020.

\bibitem[Zhou et~al.(2019)Zhou, Lan, Liu, and Yosinski]{zhou2019deconstructing}
Hattie Zhou, Janice Lan, Rosanne Liu, and Jason Yosinski.
\newblock Deconstructing lottery tickets: Zeros, signs, and the supermask.
\newblock In \emph{Advances in Neural Information Processing Systems (NeurIPS)}, 2019.

\end{thebibliography}
\bibliographystyle{iclr2024_conference}

\clearpage
\newpage
\appendix
\section{Related Works}
\label{app:sec:related_work}

\paragraph{Lottery ticket hypothesis.}
Neural network pruning techniques aim to identify parameters that can be removed from the network without affecting its performance. The simplest yet effective pruning strategy is \textit{magnitude-based pruning}, which involves removing parameters with small magnitudes~\citep{janowsky1989pruning,han2015learning}. While there is no guarantee that zeroing the weights that are close to zero will mitigate the increase in training loss~\citep{lecun1989optimal,hassibi1992second}, magnitude-based pruning can still be effective as it minimizes changes in the output of each layer~\citep{dong2017learning,park2020lookahead}. Empirical studies have demonstrated that \textit{iterative magnitude pruning (IMP)} - applying magnitude-based pruning multiple times - can effectively remove a large proportion of weights in neural networks~\citep{frankle2019lottery,renda2020comparing,frankle2020linear}. 

A large body of work tries to unravel the success of lottery tickets given the increasing interest across many applications~\citep{chen2021unified, kalibhat2021winning, girish2021lottery}. In light of existing sanity-checking methods such as score inversion and layer-wise random shuffling, IMP passes those sanity checks while other one-shot pruning methods fail, which shows the matching tickets do contain useful information regarding network weights~\citep{ma2021sanity, frankle2020pruning, su2020sanity}. \cite{zhou2019deconstructing} focus on the {\textit zeroing} out weights explaining such operation behaves resembles neural network training along with putting emphasis on preserving the sign of weights. \cite{renda2020comparing} proposes learning rate rewinding which is empirically shown to achieve superior performance compared to weight rewinding and fine-tuning. Inspired by Gordon's escape theorem, \cite{larsen2021many} theoretically relates the matching ticket to a burn-in random subspace crossing the loss sublevel set. 

\paragraph{More pruning literature.}
Lottery tickets are primarily hindered by their substantial training costs whereas there exist one-shot pruning methods which only require cost corresponding to a single cycle of IMP. As a pioneering work, \cite{lee2018snip} proposes SNIP which approximates the sensitivity of a connection, i.e., the impact of removing a single connection. GraSP~\citep{wang2020picking} suggests preserving the gradient flow as a pruning criterion, and SynFlow~\citep{tanaka2020pruning} avoids the layer-collapse phenomenon without looking at the training data. With the intention of catching up with such computationally efficient algorithms, \cite{you2019drawing} proposes early-bird tickets which significantly reduces the cost of the mask-searching step. As IMP provides so-called sparse-to-sparse training, i.e., the mask is pre-defined, some work investigates dense-to-sparse training with dynamic masks~\citep{evci2020rigging, liu2020dynamic}. \cite{liu2020dynamic} employs a trainable mask layer jointly optimizing masking threshold and weight throughout a single cycle. \cite{evci2020rigging} proposes a rigged lottery (RigL) that regrows the dead weights yet with a large gradient flow. It has been demonstrated empirically that RigL can escape local minima and discover improved basins during optimization process.

\paragraph{Loss landscape and weight averaging.}
Recent studies have shown that the loss landscape of modern deep neural networks exhibits \textit{mode-connectivity}, which means that the local minima found through stochastic optimization are connected by continuous low-loss paths~\citep{draxler2018essentially,garipov2018loss}. It motivates researchers to explore subspaces in the weight space that contain high-performing solutions~\citep{izmailov2020subspace,wortsman2021learning}. One promising approach to obtain such solutions is \textit{weight averaging}, which involves taking the average of weights across multiple points~\citep{izmailov2018averaging,wortsman2022model}. The main assumption of the weight averaging strategy is that all points are located in the same basin of attraction, which may not hold in the case of sparse networks where different sparse structures may correspond to different areas in the loss landscape. However, recent works~\citep{frankle2020linear,evci2022gradient,paul2023unmasking} suggest that this assumption may hold even for sparse networks, as there is a linear connectivity between the solutions of different sparsities obtained through iterative magnitude pruning. \cite{frankle2020linear} show that two neural networks with the same matching ticket yet trained with different SGD noises are linearly connected without a high loss barrier in between. \cite{evci2022gradient} argues that lottery tickets are biased towards the final pruned solution, i.e., winning tickets enables relearning the previous solution. \cite{paul2023unmasking} identify the existence of a low-loss curve connecting two solutions from consecutive IMP cycles, which guarantees that SGD noise cannot significantly alter the optimization path.
Recently, \cite{yin2023lottery} propose to weight-average the tickets generated from consecutive IMP iterations in order to obtain a single stronger sparse network. 
\section{Experimental Details}
\label{app:sec:details}

Our algorithm is built on Pytorch 1.10.2~\citep{Paszke_PyTorch_An_Imperative_2019}, which is available under a BSD-style license\footnote{\href{https://github.com/pytorch/pytorch/blob/main/LICENSE}{https://github.com/pytorch/pytorch/blob/main/LICENSE}}. All experiments are conducted on NVIDIA RTX 2080 and NVIDIA RTX 3090 machines.

\paragraph{Pruning.}
Throughout our experiments, we exclusively perform weight pruning on the convolutional layer, leaving the batch normalization and fully-connected layers unpruned. Additionally, we do not utilize any predefined layer-wise pruning ratio. It is worth noting that we empirically checked that including the fully-connected layers as part of the prunable parameters does not affect the results we have obtained. By default, we use the constant pruning ratio of $\alpha=0.2$ at all IMP pruning rounds, i.e., we retain 80\% of weight. For the convenience of the readers, we provide approximate sparsity values at the 3rd, 6th, 10th, 13th cycles as 50\%, 75\%, 90\%, and 95\%, respectively, while the precise sparsity values are 48.80\%, 73.79\%, 89.26\%, and 94.50\%. Unless otherwise specified, we set the number of SWAMP particles to four.

\paragraph{Baselines. }
For the fair comparison in terms of training epochs, we train all the baselines listed in \cref{main:subsec:main_results} until convergence. Merely increasing the number of training epochs does not benefit baselines; the performance of SWAMP cannot be pertained to its extensive training computations. For one-shot pruning methods~\citep{lee2018snip, wang2020picking, tanaka2020pruning}, we post-train for 300 epochs after obtaining the sparse mask, and follow the same hyperparameter setting in the original paper. 
For DST~\citep{liu2020dynamic}, we explore $\alpha \in [1e-7, 1e-5]$ for all models and datasets to match similar sparsity levels with other baselines. For RigL~\citep{evci2020rigging}, we search $\alpha \in [0.3, 0.5, 0.7]$. Here we fixed hyperparameter $\delta=100$.  
We explore $\rho \in \{0.005, 0.01, 0.05, 0.1, 0.5\}$ for SAM~\citep{na2022train} with the same hyperparameter and optimizer setting of SWAMP. 
For Lottery Pools~\citep{yin2023lottery}, we explore $\alpha \in \{0.01, 0.05, 0.1, 0.2, 0.3, 0.4, 0.5, 0.6, 0.7, 0.8, 0.9, 0.99\}$, and greedily choose among 7 candidate models w.r.t. validation accuracy. 


\paragraph{Optimization.}
We utilize the SGD optimizer with a momentum of 0.9 and a learning rate schedule that follows the cosine-decaying method. We choose the cosine-decaying schedule as it yields better results compared to the step decay schedule commonly used in lottery ticket literature.

\paragraph{CIFAR-10/100.}
To obtain a matching solution, we first train the model for 10 epochs using a constant learning rate of 0.1, a weight decay 1e-4, and a batch size of 128. Subsequently, we employ a cosine decaying learning rate schedule over 150 training epochs in each cycle of the IMP algorithm. During the final quarter of the training epochs, we collect particles for SWAMP, resulting in a total of 38 particles in a single trajectory. The learning rate for this phase is set to a constant value of 0.05.

\paragraph{Tiny-ImageNet.}
To obtain a matching solution, we first train the model for 20 epochs using a constant learning rate of 0.1, a weight decay 5e-4, and a batch size of 256. Subsequently, we employ a cosine decaying learning rate schedule over 160 training epochs in each cycle of the IMP algorithm. During the final quarter of the training epochs, we collect particles for SWAMP, resulting in a total of 40 particles in a single trajectory. The learning rate for this phase is set to a constant value of 0.05.

\paragraph{ImageNet.}
To obtain a matching solution, we first train the model for 30 epochs using a constant learning rate of 0.8, a weight decay 1e-4, and a batch size of 2048. Subsequently, we employ a cosine decaying learning rate schedule over 60 training epochs in each cycle of the IMP algorithm. During the final quarter of the training epochs, we collect particles for SWAMP, resulting in a total of 15 particles in a single trajectory. The learning rate for this phase is set to a constant value of 0.004. We establish a distributed training environment comprising eight machines, and consequently, we distribute particles across these eight machines as discussed in \cref{main:subsec:ablation}.
We also employed the \textit{learning rate rewinding}~\citep{renda2020comparing} instead of \textit{weight rewinding} in ImageNet experiments.

\section{Supplementary Results}
\label{app:sec:additional_experiments}

\subsection{Algorithms}
\label{app:sec:additional_experiments:algorithms}

\paragraph{Stochastic Weight Averaging}
\cref{table/algorithm_sgd,table/algorithm_swa} describe a detailed procedure for stochastic gradient descent~\citep[SGD;][]{robbins1951stochastic} and stochastic weight averaging~\citep[SWA;][]{izmailov2018averaging}, which are respectively denoted as $\operatorname{SGD}_{0 \rightarrow T}(\bsw_0, \xi, \calD)$ and $\operatorname{SWA}_{0 \rightarrow T}(\bsw_0, \xi, \calD)$ in the main text of the paper for notational simplicity. For SWA, we took a moving average over model copies sampled from the last 25\% of the training epochs.

\begin{algorithm}[t]
\caption{Stochastic Gradient Descent (i.e., $\operatorname{SGD}_{0 \rightarrow T}(\bsw_0, \xi, \calD)$ )}
\label{table/algorithm_sgd}
\begin{algorithmic}[1]
\Require An initial neural network parameter $\bsw_0$, training dataset $\calD$, the number of training iteration $T$, a learning rate $\eta$, a loss function $\calL$, and SGD noise $\xi$.
\Ensure Solution $\bsw_T$.
\item[]
\For{$t\in\{1,\dots,T\}$}
    \State Sample a mini-batch $\calB\subset\calD$.
    \State Update parameter $\bsw_{t} \gets \bsw_{t-1} - \eta\cdot\nabla_{\bsw}\calL(\bsw;\calB)$.
\EndFor
\end{algorithmic}
\end{algorithm}

\begin{algorithm}[t]
\caption{Stochastic Weight Averaging (i.e., $\operatorname{SWA}_{0 \rightarrow T}(\bsw_0, \xi, \calD)$ )}
\label{table/algorithm_swa}
\begin{algorithmic}[1]
\Require An initial neural network parameter $\bsw_0$, training dataset $\calD$, the number of training iteration $T$, a learning rate $\eta$, a loss function $\calL$, and SGD noise $\xi$.
\Ensure Solution $\bsw_{\text{SWA}}$.
\item[]
\State Initialize $\bsw_{\text{SWA}} \gets \boldsymbol{0}$ and $n_{\text{SWA}} \gets 0$.
\For{$t\in\{1,\dots,T\}$}
    \State Sample a mini-batch $\calB\subset\calD$.
    \State Update parameter $\bsw_{t} \gets \bsw_{t-1} - \eta\cdot\nabla_{\bsw}\calL(\bsw;\calB)$.
    \State Periodically update $\bsw_{\text{SWA}} \gets (n_{\text{SWA}}\bsw_{\text{SWA}} + \bsw_{t}) / (n+1)$ and $n_{\text{SWA}} \gets n_{\text{SWA}}+1$.
\EndFor
\end{algorithmic}
\end{algorithm}

\paragraph{Iterative Magnitude Pruning}
\begin{algorithm}[t]
\caption{Iterative Magnitude Pruning~\citep{frankle2019lottery}}
\label{table/algorithm_imp_original}
\begin{algorithmic}[1]
\Require Neural network parameter $\bsw$, pruning mask $\bsm$, training dataset $\calD$, the number of cycles for iterative magnitude pruning $C$, the number of iterations for each cycle $T$, pruning ratio $\alpha$, and SGD noise $\xi$.
\Ensure Sparse solution $\bsw_{c,T}$.
\item[]
\State Randomly initialize $\bsw_{0,0}$ and set mask $\bsm \gets \boldsymbol{1}$. \Comment{Starts from random dense weights.}
\For{$c\in\{1,\dots,C\}$}
    \State Reset $\bsw_{c,0} \gets \bsw_{0,T_0} \circ \bsm$. \Comment{Starts from the matching ticket.}
    \State Train $\bsw_{c,T} \gets \textsc{SGD}_{0 \rightarrow T}(\bsw_{c,0}, \xi_c, \calD)$. \Comment{Averages weights over trajectory.}
    \State Prune $\bsm \gets \operatorname{Prune}(\bsw_{c,T}, \alpha)$. \Comment{Updates the mask based on magnitudes.}
\EndFor
\end{algorithmic}
\end{algorithm}

\begin{algorithm}[t]
\caption{Iterative Magnitude Pruning {\color{RoyalBlue}with Rewinding}~\citep{frankle2020linear}}
\label{table/algorithm_imp}
\begin{algorithmic}[1]
\Require Neural network parameter $\bsw$, pruning mask $\bsm$, training dataset $\calD$, the number of cycles for iterative magnitude pruning $C$, the number of iterations for each cycle $T$, pruning ratio $\alpha$, SGD noise $\xi$, {\color{RoyalBlue}and the number of iteration for matching ticket $T_0$}.
\Ensure Sparse solution $\bsw_{c,T}$.
\item[]
\State Randomly initialize $\bsw_{0,0}$ and set mask $\bsm \gets \boldsymbol{1}$. \Comment{Starts from random dense weights.}
\color{RoyalBlue}
\State Train $\bsw_{0,T_0} \gets \textsc{SGD}_{0 \rightarrow T_0}(\bsw_{0,0}\circ\bsm, \xi_0, \calD)$. \Comment{Gets a matching ticket from the initialization.}
\color{black}
\For{$c\in\{1,\dots,C\}$}
    \color{RoyalBlue}
    \State Rewind $\bsw_{c,0} \gets \bsw_{0,T_0} \circ \bsm$. \Comment{Starts from the matching ticket.}
    \color{black}
    \State Train $\bsw_{c,T} \gets \textsc{SGD}_{0 \rightarrow T}(\bsw_{c,0}, \xi_c, \calD)$. \Comment{Averages weights over trajectory.}
    \State Prune $\bsm \gets \operatorname{Prune}(\bsw_{c,T}, \alpha)$. \Comment{Updates the mask based on magnitudes.}
\EndFor
\end{algorithmic}
\end{algorithm}

\cref{table/algorithm_imp_original} and \cref{table/algorithm_imp} respectively provide a detailed procedure for the vanilla iterative magnitude pruning~\citep[IMP;][]{frankle2019lottery} and IMP with weight rewinding~\citep[IMP-WR;][]{frankle2020linear}. While IMP rewinds the network to the initialization after each cycle, IMP-WR rewinds the weights to the early epoch of the training process. $\operatorname{Prune}(\bsw,\alpha)$ returns a pruning mask where $(1-\alpha)$ of the remaining parameters are pruned based on their magnitudes.

\subsection{Extension to language tasks}
\label{app:sec:additional_experiments:nlp}

\cref{table/main_roberta} verified the capability of SWAMP to language tasks. Specifically, we fine-tuned the {\it RoBERTa-Base} model~\citep{liu2019roberta} on two subtasks from GLUE benchmark~\citep{wang2018glue}: Microsoft Research Paraphrase Corpus~\citep[MRPC;][]{dolan2005automatically} and Recognizing Textual Entailment~\citep[RTE;][]{dagan2006pascal,bar2006second,giampiccolo2007third,bentivogli2009fifth}. We mainly followed the experimental configuration outlined in the work of \cite{liu2019roberta}, which includes specifications such as a token limit of 512, and utilization of the Adam optimizer~\citep{kingma2015adam} with first-moment coefficient set to 0.9 and second-moment coefficient set to 0.98.

\begin{figure}[t]
    \centering
    \includegraphics[width=0.7\linewidth]{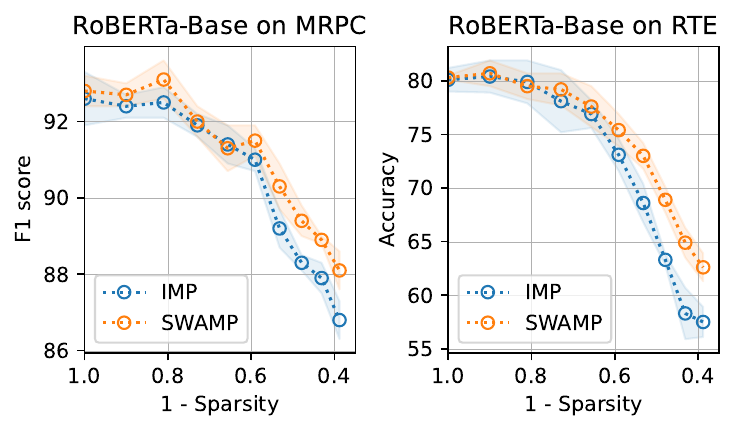}
    \caption{Results for RoBERT on MRPC and RTE. Reported values are averaged over three random seeds.}
    \label{figure/nlp}
\end{figure}

To obtain a matching solution, we first train the model for 5 epochs using a constant learning rate of 1e-05. Subsequently, we employ a linear decaying learning rate schdule over 10 training epochs in each cycle of the IMP algorithm. For SWAMP. we collect particles distributed across eight machines during the final half of the training epochs. \cref{figure/nlp} summarizes the results for each IMP cycle, where we used the pruning ratio of $\alpha=0.9$. We can readily find out that SWAMP outperforms the IMP baseline by a noticeable margin on both datasets.

\subsection{Extension to dynamic pruning methods}
\label{app:sec:additional_experiments:rigl}

\begin{table}[t]
    \centering
    \caption{Further comparison between (a) RigL and (b) RigL + SWAMP on CIFAR-10. Reported values are averaged over  three random seeds.}
    \label{table/rigl}
    \resizebox{0.5\linewidth}{!}{\begin{tabular}{ *{4}{c} }
    \toprule
    & \multicolumn{3}{c}{Sparsity} \\
    \cmidrule(lr){2-4}
    WRN28x2 & 75\% & 90\% & 95\% \\
    \midrule
    (a) & 93.09$\spm{0.13}$ & 91.81$\spm{0.18}$ & 90.73$\spm{0.16}$ \\
    (b) & \BF{93.41}$\spm{0.04}$ & \BF{92.78}$\spm{0.10}$ & \BF{92.40}$\spm{0.46}$ \\
    \midrule
    \midrule
    & \multicolumn{3}{c}{Sparsity}\\
    \cmidrule(lr){2-4}
    VGG-13 & 75\% & 90\% & 95\% \\
    \midrule
    (a) & 93.18$\spm{0.04}$ & 92.46$\spm{0.16}$ & 91.45$\spm{0.16}$ \\
    (b) & \BF{93.53}$\spm{0.08}$ & \BF{93.01}$\spm{0.16}$ & \BF{93.01}$\spm{0.23}$ \\
    \bottomrule
    \end{tabular}}
\end{table}

In this paper, we extend IMP to SWAMP leveraging particle averaging heavily inspired by \cite{paul2023unmasking}. Although the primary focus of our paper is the loss landscape analysis of IMP, one can readily apply SWAMP to dynamic pruning methods as well. Here we further extend our algorithm to encompass dense-to-sparse training, offering a means to mitigate the intensive training costs associated with IMP which includes multiple rounds of retraining. Our approach enhances RigL~\citep{evci2020rigging} by introducing multiple particles during the final training phase, integrated after completing 75\% of the total training epochs. The results presented in \cref{table/rigl} demonstrate that our method gracefully merges with RigL, incurring only marginal training overhead. This result shows that SWAMP can be naturally applied to dynamic pruning methods.

\subsection{Further ablation studies}
\label{app:sec:additional_experiments:ablation}

\begin{table}[t]
\caption{Ablation study on the number of SWAMP particles. Performance improves with an increase in the number of particles. Reported values are classification accuracy averaged over three random seeds, and the best and second-best results are boldfaced and underlined, respectively.}
\label{table/app_number_of_particles}
\setlength{\tabcolsep}{1.0em}
\centering
\resizebox{1.0\linewidth}{!}{\begin{tabular}{cccccc}
    \toprule
    & & \multicolumn{4}{c}{Sparsity}\\
    \cmidrule(lr){3-6}
    & \# particles & 50\% & 75\% & 90\% & 95\%  \\
    \midrule
    \multirow{4}{*}{\Centerstack{CIFAR-10\\(WRN-28-2)}}
        & 1 & 94.62 $\pm$ 0.06 & 
            94.67 $\pm$ 0.06 & 
            94.35 $\pm$ 0.06 & 
            93.97 $\pm$ 0.10 \\
        & 2 & 94.57 $\pm$ 0.04 & 
            94.59 $\pm$ 0.07 & 
            94.38 $\pm$ 0.17 & 
            94.05 $\pm$ 0.16 \\
        & 4 & \underline{94.74} $\pm$ 0.04 & 
            \underline{94.88} $\pm$ 0.09 & 
            \underline{94.73} $\pm$ 0.10 & 
            \bf{94.23 $\pm$ 0.11} \\
        & 8 & \bf{94.80 $\pm$ 0.04} & 
            \bf{94.90 $\pm$ 0.09} & 
            \bf{94.74 $\pm$ 0.10} & 
            \underline{94.21} $\pm$ 0.24 \\
    \midrule   
    \multirow{4}{*}{\Centerstack{CIFAR-10\\(VGG-13)}}
        & 1 & 94.07 $\pm$ 0.03 & 
        94.15 $\pm$ 0.09 & 
        94.12 $\pm$ 0.22 & 
        93.99 $\pm$ 0.17 \\
        & 2 & 94.03 $\pm$ 0.10 & 
        93.93 $\pm$ 0.16 & 
        94.01 $\pm$ 0.09 & 
        94.03 $\pm$ 0.07 \\
        & 4 & \underline{94.14} $\pm$ 0.08 & 
        \underline{94.39} $\pm$ 0.15 & 
        \underline{94.40} $\pm$ 0.16 & 
        \underline{94.34} $\pm$ 0.11  \\
        & 8 & \bf{94.24 $\pm$ 0.18} & 
        \bf{94.41 $\pm$ 0.09} & 
        \bf{94.54 $\pm$ 0.09} & 
        \bf{94.41 $\pm$ 0.15} \\
    \bottomrule
\end{tabular}}
\end{table}

\begin{table}[t]
\caption{Ablation study to validate SWAMP's efficacy in two aspects: (i) mask generation and (ii) sparse training. Except for earlier iterations, i.e., low sparsity regime, our method excels in both areas relative to vanilla SGD optimization. Reported values are classification accuracy averaged over three random seeds, and the best results are boldfaced.}
\label{table:app_mask_and_training}
\setlength{\tabcolsep}{0.9em}
\centering
\resizebox{0.9\linewidth}{!}{\begin{tabular}{cllcccc}
    \toprule
     & & & \multicolumn{4}{c}{Sparsity} \\
     \cmidrule(lr){4-7}
     & Mask & Training & 50\% & 75\% & 90\% & 95\%  \\
     \midrule
    \multirow{4}{*}{\Centerstack{CIFAR-10\\(VGG-13)}}
        & SGD   & SGD & 93.34$\spm{0.17}$ & 93.53$\spm{0.15}$ & 93.58$\spm{0.03}$ & 93.62$\spm{0.11}$\\ 
        & SGD   & SWAMP & 94.01$\spm{0.02}$ & 94.30$\spm{0.01}$ & 94.35$\spm{0.08}$ & 94.18$\spm{0.13}$\\ 
        \cmidrule(lr){2-7}
        & SWAMP & SGD & 93.55$\spm{0.03}$ & 93.84$\spm{0.10}$ & 94.07$\spm{0.07}$ & 93.98$\spm{0.07}$\\ 
        & SWAMP & SWAMP & {\bf 94.14}$\spm{0.08}$ & {\bf 94.39}$\spm{0.15}$ & {\bf 94.40}$\spm{0.16}$ & {\bf 94.34}$\spm{0.11}$\\ 
    \bottomrule
\end{tabular}}
\end{table}

\begin{table}[t]
\centering  
\caption{Ablation study on the impact of the two main components of SWAMP; averaging multiple particles (denoted by Multi) and averaging learning trajectory (denoted by SWA). Our findings indicate that the best performance is achieved when both techniques are employed. Reported classification accuracies are averaged over three random seeds, and the best and second-best results are boldfaced and underlined, respectively.}
\label{table:app_sgd_swa}
\setlength{\tabcolsep}{0.9em}
\resizebox{0.9\linewidth}{!}{\begin{tabular}{ccccccc}
    \toprule
    & & & \multicolumn{4}{c}{Sparsity} \\
    \cmidrule(lr){4-7}
    & Multi & SWA & 50\% & 75\% & 90\% & 95\%  \\
    \midrule
    \multirow{4}{*}{\Centerstack{CIFAR-10\\(VGG-13)}}
        & \cmark & \cmark 
        & {\bf 94.14}$\spm{0.08}$ & 
        {\bf 94.39}$\spm{0.15}$ & 
        {\bf 94.40}$\spm{0.16}$ & 
        {\bf 94.34}$\spm{0.11}$ \\ 
        & \cmark & \xmark 
        & 93.96$\spm{0.03}$ & 
        94.09$\spm{0.09}$ & 
        94.08$\spm{0.18}$ & 
        \underline{94.05}$\spm{0.11}$ \\
        & \xmark & \cmark 
        & \underline{94.07}$\spm{0.03}$ & 
        \underline{94.15}$\spm{0.09}$ & 
        \underline{94.12}$\spm{0.22}$ & 
        93.99$\spm{0.17}$ \\ 
        & \xmark & \xmark 
        & 93.34$\spm{0.17}$ & 
        93.53$\spm{0.15}$ & 
        93.58$\spm{0.03}$ & 
        93.62$\spm{0.11}$ \\ 
    \bottomrule
\end{tabular}}
\end{table}

As described in \cref{main:subsec:swamp_algorithm}, we conduct an empirical analysis to investigate the correlation between the number of particles and the performance of SWAMP. \cref{table/app_number_of_particles} presents our findings, demonstrating a strong positive correlation between the generalization performance of SWAMP and the number of individual particles. These results indicate that maximizing the number of particles used in SWAMP can significantly enhance model performance. Additionally, they provide evidence supporting the mention made in \cref{main:subsec:swamp_algorithm} that both SWA and particle averaging contribute independently and complementarily to the overall effectiveness of the SWAMP algorithm. 
We also provide additional ablation results with VGG networks in \cref{table:app_mask_and_training} and \cref{table:app_sgd_swa}.

\subsection{Further results on Tiny-ImageNet}
\label{app:sec:additional_experiments:imagenet}

\begin{figure}[t]
\centering
\includegraphics[width=\linewidth]{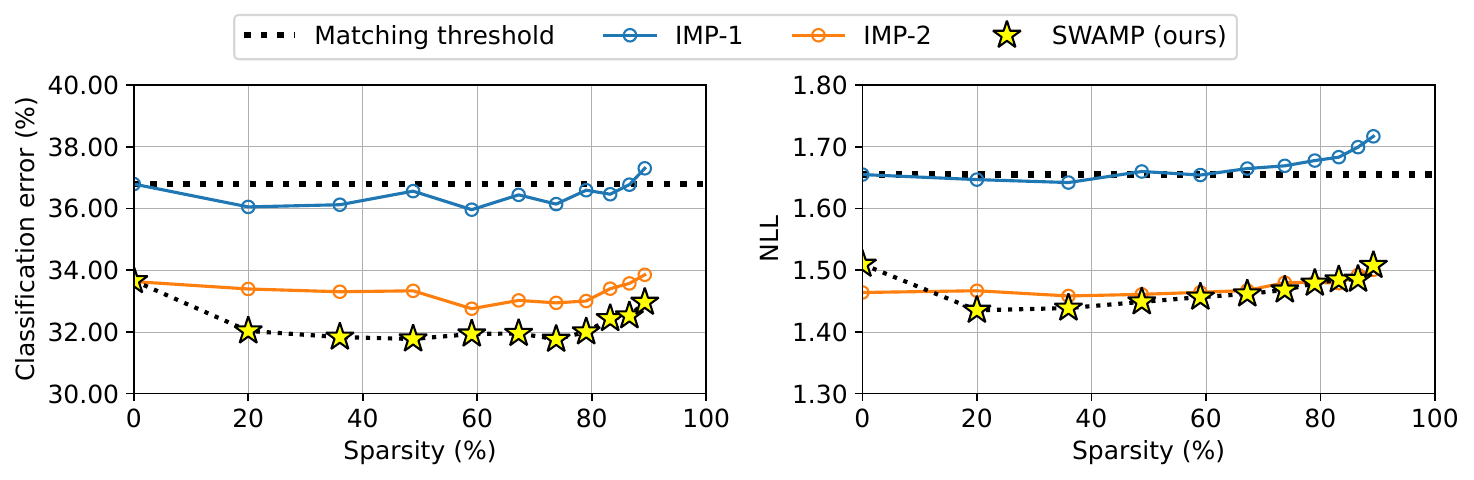}
\caption{Classification error (left) and negative log-likelihood as a function of the sparsity. SWAMP achieves remarkable performance comparable to an ensemble of IMP solutions while maintaining the same inference cost as a single model. Here $n$ in the notation IMP-$n$ indicates the number of IMP ensemble members. Notably, SWAMP demonstrates \textit{matching} performance even at extremely sparse levels, unlike IMP. The results are presented for ResNet-18 on Tiny-ImageNet.}
\label{figure/sparsity_tiny_appendix}
\end{figure}

\cref{figure/sparsity_tiny_appendix} depicts the results on Tiny-ImageNet. In the case of ResNet-18 on Tiny-ImageNet, SWAMP outperforms the ensemble performance of two sparse networks. 

\subsection{Sparsity plots}
\label{app:sec:additional_experiments:sparsity}

\begin{figure}[t]
    \centering
    
    \begin{subfigure}{\textwidth}
    \centering
    \includegraphics[width=\linewidth]{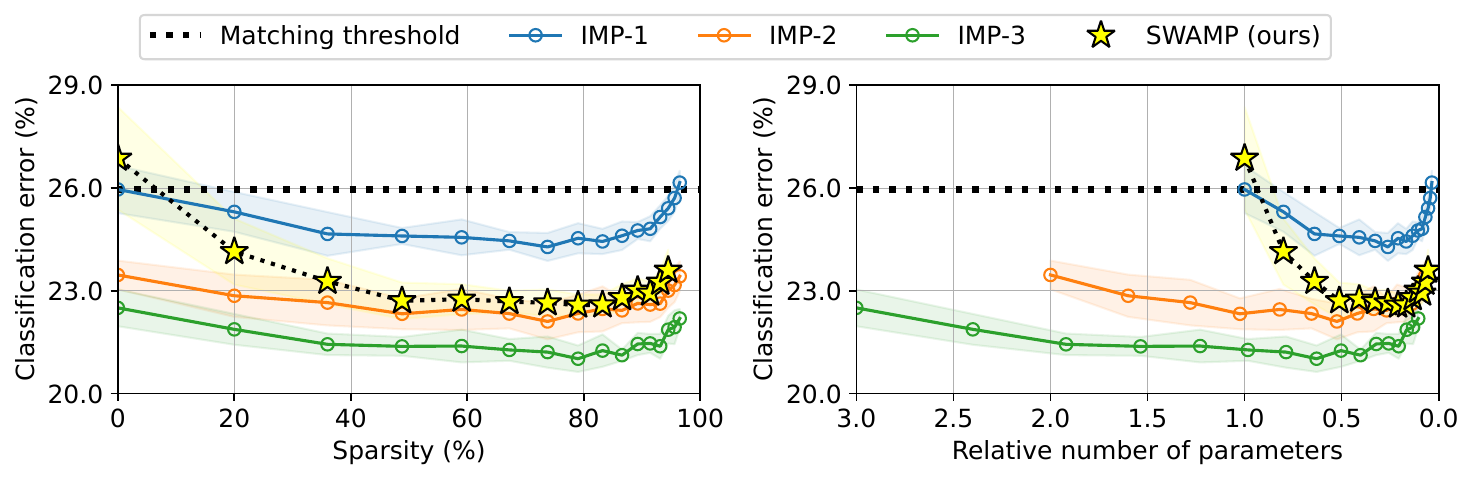}
    \caption{WRN-32-4 on CIFAR-100}
    \end{subfigure}
    \vspace{0.5em}
    
    \begin{subfigure}{\textwidth}
    \centering
    \includegraphics[width=\linewidth]{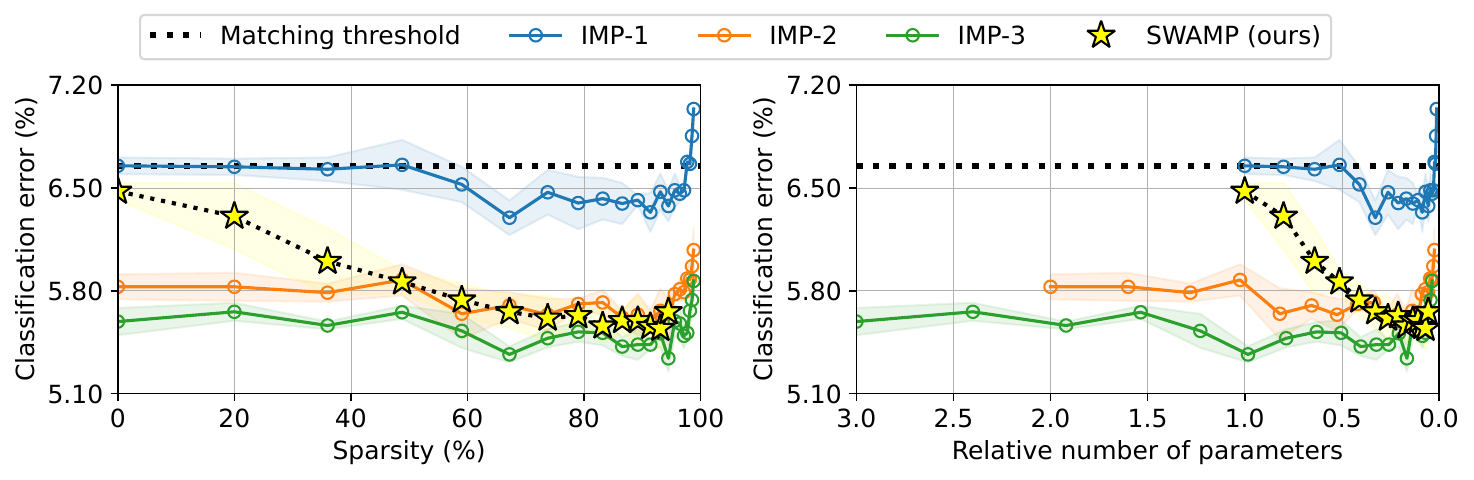}
    \caption{VGG-13 on CIFAR-10}
    \end{subfigure}
    \vspace{0.5em}
    
    \begin{subfigure}{\textwidth}
    \centering
    \includegraphics[width=\linewidth]{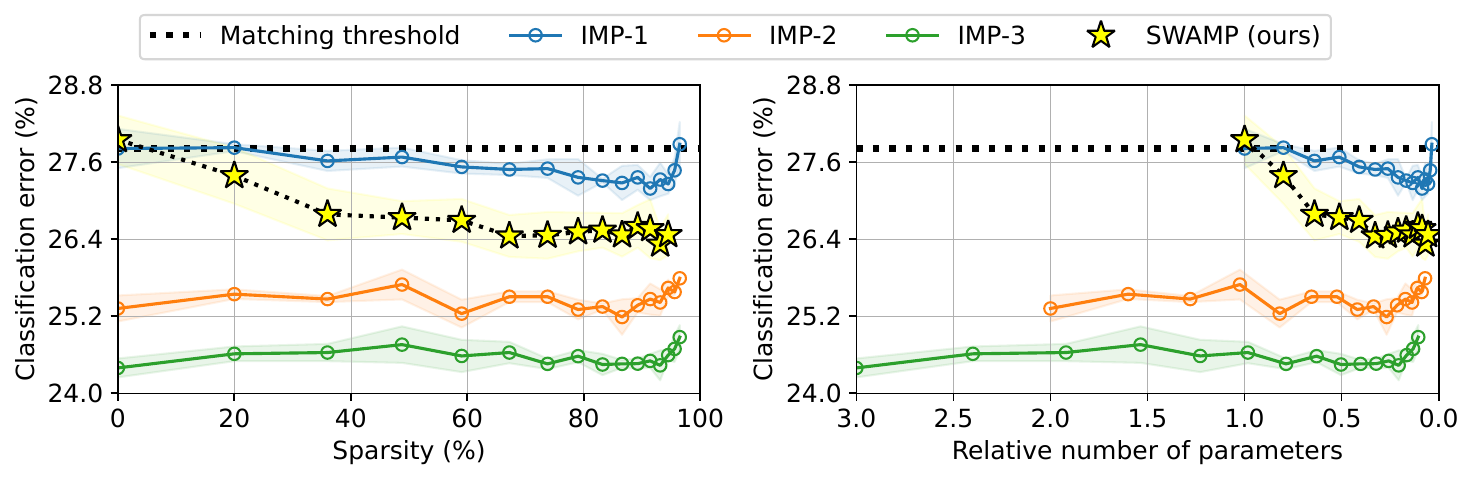}
    \caption{VGG-16 on CIFAR-100}
    \end{subfigure}
    
    \caption{Supplementary plots for \cref{figure/sparsity}. Classification error as a function of the sparsity (left) and the relative number of parameters (right). Again, SWAMP demonstrates \textit{matching} performance even at extremely sparse levels, unlike IMP.}
    \label{figure/sparsity_appendix}
\end{figure}

\begin{table}[t]
\caption{Classification accuracy on VGG networks. SWAMP outperforms all the baselines across varying sparsities, including one-shot and dynamic pruning approaches. Reported values are averaged over three random seeds, and the best and second-best results are boldfaced and underlined, respectively.}
\label{table/main_cifar100}
\setlength{\tabcolsep}{0.9em}
\centering
\resizebox{1.00\linewidth}{!}{\begin{tabular}{clcccc}
    \toprule
    &        & \multicolumn{4}{c}{Sparsity}\\
    \cmidrule(lr){3-6}
    & Method & 50\% & 75\% & 90\% & 95\%  \\
    \midrule
    \multirow{9}{*}{\Centerstack{CIFAR-10\\(VGG-13)}}
        & SNIP              & 92.85 $\pm$ 0.19 & 92.59 $\pm$ 0.22 & 91.30 $\pm$ 0.26 & 90.34 $\pm$ 0.25 \\
        & SynFlow           & 93.01 $\pm$ 0.27 & 93.09 $\pm$ 0.27 & 92.74 $\pm$ 0.40 & 91.54 $\pm$ 0.21 \\
        & GraSP             & 92.20 $\pm$ 0.06 & 91.94 $\pm$ 0.27 & 91.16 $\pm$ 0.11 & 90.68 $\pm$ 0.08 \\
        & RigL              & 93.56 $\pm$ 0.18 & 93.18 $\pm$ 0.04 & 92.46 $\pm$ 0.16 & 91.45 $\pm$ 0.16 \\
        & DST               & \underline{93.93} $\pm$ 0.20 & 93.90 $\pm$ 0.06 & 93.75 $\pm$ 0.02 & 93.38 $\pm$ 0.06 \\
        & IMP               & 93.34 $\pm$ 0.17 & 93.53 $\pm$ 0.15 & 93.58 $\pm$ 0.03 & 93.62 $\pm$ 0.11 \\
        & IMP + SAM         & 93.73 $\pm$ 0.12 & \underline{93.96} $\pm$ 0.19 & \underline{94.01} $\pm$ 0.12 & \underline{93.89} $\pm$ 0.13 \\
        \cmidrule(lr){2-6}
        & \bf{SWAMP (ours)} & \bf{94.14} $\pm$ \bf{0.08} & \bf{94.39} $\pm$ \bf{0.15} & \bf{94.40} $\pm$ \bf{0.16} & \bf{94.34} $\pm$ \bf{0.11} \\
    \midrule   
    \multirow{9}{*}{\Centerstack{CIFAR-100\\(VGG-16)}}
        & SNIP              & 71.21 $\pm$ 0.21 & 70.41 $\pm$ 0.31 & 67.96 $\pm$ 0.15 & 66.24 $\pm$ 0.40 \\
        & SynFlow           & 71.52 $\pm$ 0.05 & 71.31 $\pm$ 0.13 & 71.00 $\pm$ 0.22 & 67.18 $\pm$ 0.24 \\
        & GraSP             & 69.08 $\pm$ 0.25 & 67.26 $\pm$ 0.06 & 65.25 $\pm$ 0.38 & 63.50 $\pm$ 0.09 \\
        & RigL              & 72.22 $\pm$ 0.43 & 71.38 $\pm$ 0.07 & 69.59 $\pm$ 0.41 & 67.44 $\pm$ 0.05 \\
        & DST               & \underline{72.93} $\pm$ 0.16 & \underline{72.87} $\pm$ 0.13 & 72.45 $\pm$ 0.10 & 71.52 $\pm$ 0.12 \\
        & IMP               & 72.32 $\pm$ 0.15 & 72.50 $\pm$ 0.15 & \underline{72.64} $\pm$ 0.19 & \underline{72.74} $\pm$ 0.15 \\
        & IMP + SAM         & 72.11 $\pm$ 0.45 & 72.27 $\pm$ 0.12 & 72.56 $\pm$ 0.19 & 72.50 $\pm$ 0.04 \\
        \cmidrule(lr){2-6}
        & \bf{SWAMP (ours)} & \bf{73.27} $\pm$ \bf{0.26} & \bf{73.54} $\pm$ \bf{0.36} & \bf{73.40} $\pm$ \bf{0.33} & \bf{73.53} $\pm$ \bf{0.32} \\
    \bottomrule
\end{tabular}}
\end{table}

\begin{table}[t]
\caption{Negative log-likelihood on CIFAR-10. SWAMP outperforms all the baselines across varying sparsities, including one-shot and dynamic pruning approaches. Reported values are averaged over three random seeds, and the best and second-best results are boldfaced and underlined, respectively.}
\label{table/appen_cifar10_nll}
\setlength{\tabcolsep}{0.9em}
\centering
\resizebox{1.00\linewidth}{!}{\begin{tabular}{clcccc}
    \toprule
    &        & \multicolumn{4}{c}{Sparsity}\\
    \cmidrule(lr){3-6}
    & Method & 50\% & 75\% & 90\% & 95\% \\
    \midrule
    \multirow{9}{*}{\Centerstack{CIFAR-10\\(WRN-28-2)}}
        & SNIP              & 0.227 $\pm$ 0.012 & 0.242 $\pm$ 0.005 & 0.282 $\pm$ 0.003 & 0.327 $\pm$ 0.007 \\
        & SynFlow           & 0.219 $\pm$ 0.007 & 0.223 $\pm$ 0.004 & 0.280 $\pm$ 0.006 & 0.316 $\pm$ 0.003 \\
        & GraSP             & 0.243 $\pm$ 0.002 & 0.281 $\pm$ 0.009 & 0.308 $\pm$ 0.017 & 0.338 $\pm$ 0.004 \\
        & RigL              & 0.206 $\pm$ 0.010 & 0.231 $\pm$ 0.005 & 0.261 $\pm$ 0.009 & 0.288 $\pm$ 0.003 \\
        & DST               & 0.189 $\pm$ 0.009 & 0.214 $\pm$ 0.008 & 0.241 $\pm$ 0.010 & 0.251 $\pm$ 0.008 \\
        & IMP               & 0.198 $\pm$ 0.001 & 0.198 $\pm$ 0.002 & 0.197 $\pm$ 0.004 & \underline{0.210} $\pm$ 0.003 \\
        & IMP + SAM         & \underline{0.182} $\pm$ 0.009 & \underline{0.180} $\pm$ 0.012 & \underline{0.187} $\pm$ 0.011 & 0.237 $\pm$ 0.053 \\
        \cmidrule(lr){2-6}
        & \bf{SWAMP (ours)} & \bf{0.158 $\pm$ 0.002} & 
                            \bf{0.156 $\pm$ 0.002} & 
                            \bf{0.162 $\pm$ 0.002} & 
                            \bf{0.177 $\pm$ 0.002} \\
    \midrule   
    \multirow{9}{*}{\Centerstack{CIFAR-10\\(VGG-13)}}
        & SNIP              & 0.234 $\pm$ 0.003 & 0.244 $\pm$ 0.008 & 0.269 $\pm$ 0.003 & 0.291 $\pm$ 0.008 \\
        & SynFlow           & 0.233 $\pm$ 0.006 & 0.224 $\pm$ 0.008 & 0.230 $\pm$ 0.007 & 0.261 $\pm$ 0.002 \\
        & GraSP             & 0.252 $\pm$ 0.002 & 0.257 $\pm$ 0.006 & 0.271 $\pm$ 0.006 & 0.282 $\pm$ 0.001 \\
        & RigL              & 0.216 $\pm$ 0.002 & 0.230 $\pm$ 0.006 & 0.246 $\pm$ 0.007 & 0.270 $\pm$ 0.002 \\
        & DST               & 0.210 $\pm$ 0.002 & 0.208 $\pm$ 0.004 & 0.213 $\pm$ 0.002 & 0.220 $\pm$ 0.002 \\
        & IMP               & 0.220 $\pm$ 0.006 & 
                            0.214 $\pm$ 0.004 & 
                            0.211 $\pm$ 0.002 & 
                            0.210 $\pm$ 0.004 \\
        & IMP + SAM         & \underline{0.195} $\pm$ 0.001 & \underline{0.191} $\pm$ 0.003 & \underline{0.190} $\pm$ 0.003 & \underline{0.192} $\pm$ 0.006 \\
        \cmidrule(lr){2-6}
        & \bf{SWAMP (ours)} & \bf{0.185 $\pm$ 0.002} & 
                            \bf{0.179 $\pm$ 0.002} & 
                            \bf{0.177 $\pm$ 0.004} & 
                            \bf{0.178 $\pm$ 0.001} \\
    \bottomrule
\end{tabular}}
\end{table}

\begin{table}[t]
\caption{Negative log-likelihood on CIFAR-100. SWAMP outperforms all the baselines across varying sparsities, including one-shot and dynamic pruning approaches. Reported values are averaged over three random seeds, and the best and second-best results are boldfaced and underlined, respectively.}
\label{table/appen_cifar100_nll}
\setlength{\tabcolsep}{0.9em}
\centering
\resizebox{1.00\linewidth}{!}{\begin{tabular}{clcccc}
    \toprule
    &        & \multicolumn{4}{c}{Sparsity}\\
    \cmidrule(lr){3-6}
    & Method & 50\% & 75\% & 90\% & 95\%  \\
    \midrule
    \multirow{9}{*}{\Centerstack{CIFAR-100\\(WRN-32-4)}}
        & SNIP              & 1.067 $\pm$ 0.010 & 1.064 $\pm$ 0.022 & 1.121 $\pm$ 0.036 & 1.17 $\pm$ 0.007 \\
        & SynFlow           & 0.964 $\pm$ 0.006 & 1.000 $\pm$ 0.019 & 0.994 $\pm$ 0.009 & 1.009 $\pm$ 0.041 \\
        & GraSP             & 1.112 $\pm$ 0.043 & 1.198 $\pm$ 0.028 & 1.235 $\pm$ 0.014 & 1.252 $\pm$ 0.036 \\
        & RigL              & 1.005 $\pm$ 0.012 & 1.047 $\pm$ 0.016 & 1.105 $\pm$ 0.029 & 1.125 $\pm$ 0.026 \\
        & DST               & 1.012 $\pm$ 0.036 & 1.029 $\pm$ 0.028 & 1.235 $\pm$ 0.014 & 1.252 $\pm$ 0.036 \\
        & IMP               & 0.974 $\pm$ 0.014 & 
                            0.973 $\pm$ 0.007 & 
                            1.001 $\pm$ 0.014 & 
                            1.010 $\pm$ 0.011 \\
        & IMP + SAM         & \underline{0.924} $\pm$ 0.031 & \underline{0.918} $\pm$ 0.028 & \underline{0.934} $\pm$ 0.024 & \underline{0.953} $\pm$ 0.023 \\
        \cmidrule(lr){2-6}
        & \bf{SWAMP (ours)} & \bf{0.847 $\pm$ 0.020} & 
                            \bf{0.851 $\pm$ 0.016} & 
                            \bf{0.868 $\pm$ 0.016} & 
                            \bf{0.893 $\pm$ 0.020}  \\
    \midrule   
    \multirow{9}{*}{\Centerstack{CIFAR-100\\(VGG-16)}}
        & SNIP              & 1.135 $\pm$ 0.015 & 1.150 $\pm$ 0.008 & 1.216 $\pm$ 0.012 & 1.259 $\pm$ 0.025 \\
        & SynFlow           & 1.123 $\pm$ 0.005 & 1.115 $\pm$ 0.006 & 1.092 $\pm$ 0.010 & 1.203 $\pm$ 0.004 \\
        & GraSP             & 1.219 $\pm$ 0.003 & 1.242 $\pm$ 0.013 & 1.311 $\pm$ 0.016 & 1.351 $\pm$ 0.005 \\
        & RigL              & 1.121 $\pm$ 0.007 & 1.142 $\pm$ 0.007 & 1.120 $\pm$ 0.016 & 1.238 $\pm$ 0.004 \\
        & DST               & 1.110 $\pm$ 0.007 & 1.121 $\pm$ 0.004 & 1.138 $\pm$ 0.004 & 1.146 $\pm$ 0.014 \\
        & IMP               & 1.105 $\pm$ 0.006 & 
                            1.096 $\pm$ 0.009 & 
                            1.090 $\pm$ 0.004 & 
                            1.096 $\pm$ 0.002 \\
        & IMP + SAM         & \underline{1.075} $\pm$ 0.014 & \underline{1.067} $\pm$ 0.010 & \underline{1.071} $\pm$ 0.004 & \underline{1.068} $\pm$ 0.007 \\
        \cmidrule(lr){2-6}
        & \bf{SWAMP (ours)} & \bf{1.023 $\pm$ 0.016} & 
                            \bf{1.029 $\pm$ 0.015} & 
                            \bf{1.033 $\pm$ 0.016} & 
                            \bf{1.029 $\pm$ 0.016}\\
    \bottomrule
\end{tabular}}
\end{table}

We provide supplementary results in \cref{figure/sparsity_appendix}, which clearly indicate that SWAMP performs better than IMP in terms of overall performance. Furthermore, we report the negative log-likelihood (NLL) values in \cref{table/appen_cifar10_nll,table/appen_cifar100_nll}, where we employed temperature scaling~\citep{guo2017calibration} for better in-domain uncertainty evaluation, as discussed in \citet{ashukha2020pitfalls}. Notably, SWAMP consistently outperforms other baselines in terms of NLL as well.


\subsection{Flatness of local minima}
\label{app:sec:additional_experiments:hessian}

\begin{table}[t]
\caption{Trace of Hessian $\text{Tr}(\bH)$ evaluated across training data. In most cases, SWAMP with multiple particles exhibits smaller trace value, i.e., finds flatter minima, compared to others. Reported values are averaged over three random seeds, and the best and second-best results are boldfaced and underlined, respectively.}
\label{table/c100_hessian_trace}
\centering  
\resizebox{1.0\linewidth}{!}{\begin{tabular}{clcccc}
    \toprule
    & & \multicolumn{4}{c}{Sparsity}\\
    \cmidrule(lr){3-6}
    & Training & 20\% & 50\% & 75\% & 90\%  \\
    \midrule
    \multirow{3}{*}{\Centerstack{CIFAR-100\\(WRN-32-4)}}
        & SGD           
        & \phantom{}16394.8 $\spm{\phantom{0}3262.1}$ 
        & \phantom{}20335.2 $\spm{\phantom{0}4875.3}$ 
        & \phantom{}29337.7 $\spm{\phantom{}10887.7}$ 
        & \phantom{}25504.9 $\spm{\phantom{0}4370.0}$ \\ 
        & SWAMP ($N=1$) 
        & \phantom{0}\underline{2586.9} $\spm{\phantom{000}10.7}$ 
        & \phantom{0}\underline{2669.4} $\spm{\phantom{000}21.9}$ 
        & \phantom{0}\underline{3079.8} $\spm{\phantom{00}150.8}$ 
        & \phantom{0}\underline{3159.3} $\spm{\phantom{00}147.3}$ \\ 
        & SWAMP ($N=4$) 
        & \phantom{0}{\bf{2457.8}} $\spm{\phantom{00}161.8}$ 
        & \phantom{0}{\bf{2556.1}} $\spm{\phantom{000}86.9}$ 
        & \phantom{0}{\bf{2896.7}} $\spm{\phantom{00}201.0}$ 
        & \phantom{0}{\bf{2968.5}} $\spm{\phantom{000}79.4}$ \\ 
    \midrule
    \multirow{3}{*}{\Centerstack{CIFAR-100\\(VGG-16)}}
        & SGD           
        & \phantom{0}4776.8 $\spm{\phantom{0}80.5}$ 
        & \phantom{0}4842.5 $\spm{\phantom{}108.0}$ 
        & \phantom{0}5033.1 $\spm{\phantom{}336.3}$ 
        & \phantom{0}5141.7 $\spm{\phantom{0}20.5}$ \\ 
        & SWAMP ($N=1$) 
        & \phantom{0}{\underline{2072.4}} $\spm{\phantom{0}34.5}$ 
        & \phantom{0}\bf{2189.2} $\spm{\phantom{0}73.9}$ 
        & \phantom{0}\underline{2472.8} $\spm{\phantom{0}93.1}$ 
        & \phantom{0}\underline{2570.1} $\spm{\phantom{0}88.8}$ \\ 
        & SWAMP ($N=4$) 
        & \phantom{0}{\bf 1995.9} $\spm{\phantom{0}39.6}$ 
        & \phantom{0}{\underline{2201.1}} $\spm{\phantom{0}85.5}$ 
        & \phantom{0}{\bf{2341.6}} $\spm{\phantom{}109.6}$ 
        & \phantom{0}{\bf{2423.8}} $\spm{\phantom{0}48.6}$ \\ 
    \bottomrule
\end{tabular}}
\end{table}

We further provide the results on the trace of Hessian for CIFAR-100 in \cref{table/c100_hessian_trace} to show SWAMP with multiple particles finds flatter local minima than IMP leading to a better generalization performance. 

\subsection{Visualization of loss surfaces}
\label{app:sec:additional_experiments:loss_surface}

\begin{figure}[t]
\centering
\includegraphics[width=0.3\linewidth]{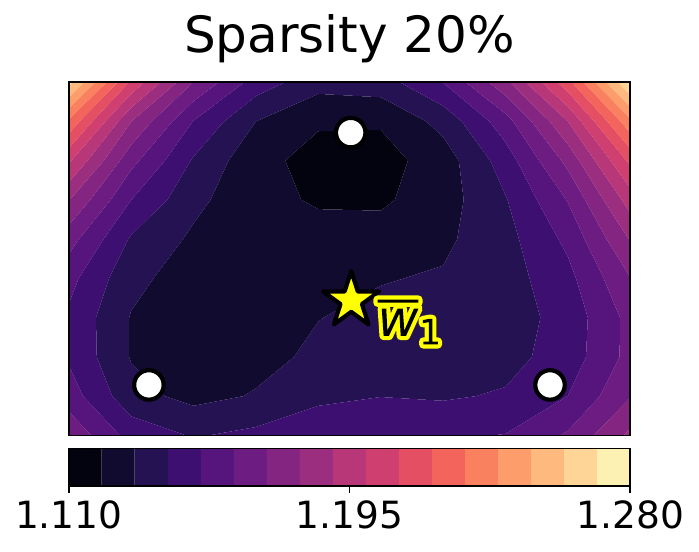}
\includegraphics[width=0.3\linewidth]{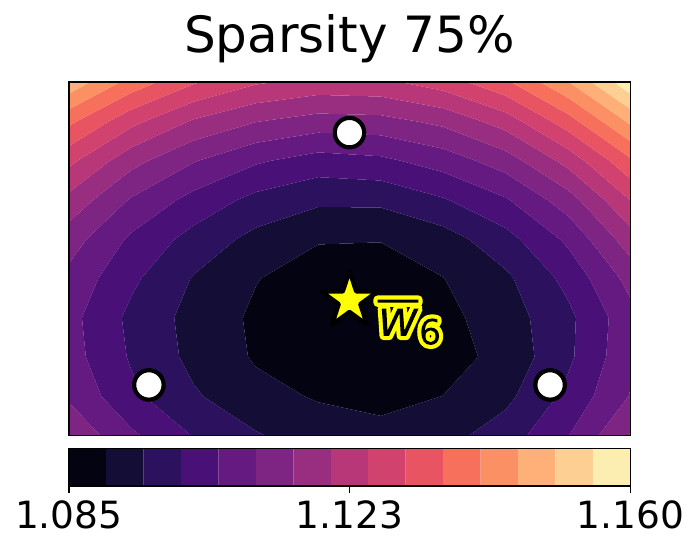}
\includegraphics[width=0.3\linewidth]{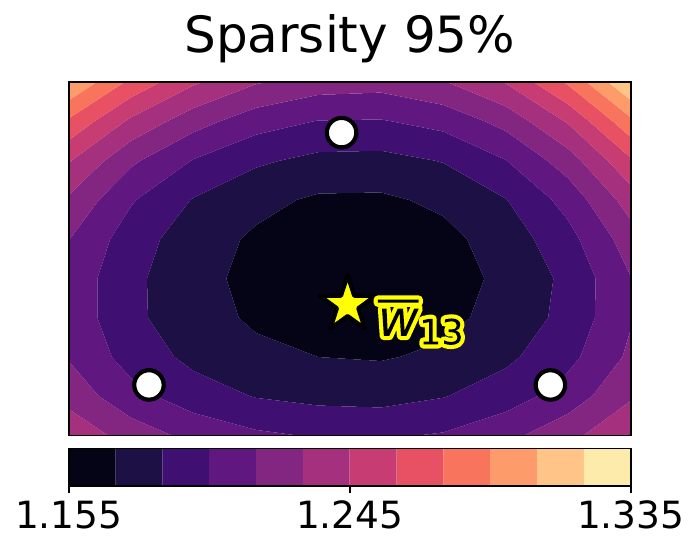}
\caption{Visualization of loss surfaces as a function of neural network weights in a two-dimensional subspace, spanned by three particles (marked as white circles). The averaged weight $\bsw_c$ (marked by a yellow star) is observed not to be positioned in the flat region of the surface during the earlier stages of IMP (left; Sparsity 20\%). However, as the sparsity increased, the weight averaging technique effectively captures the flat region of the surface. The results are presented for WRN-32-4 on the test split of CIFAR-100.}
\label{figure/loss_surface_appendix}
\end{figure}

We provide additional loss landscape visualization plots for CIFAR-100 using WRN-32-4 model in \cref{figure/loss_surface_appendix}. We can observe a similar trend that SWAMP averages better as network sparsity grows.

\subsection{Particle-wise performance}
\label{app:sec:additional_experiments:particle}

\begin{figure}[t]
\centering
\includegraphics[width=0.3\linewidth]{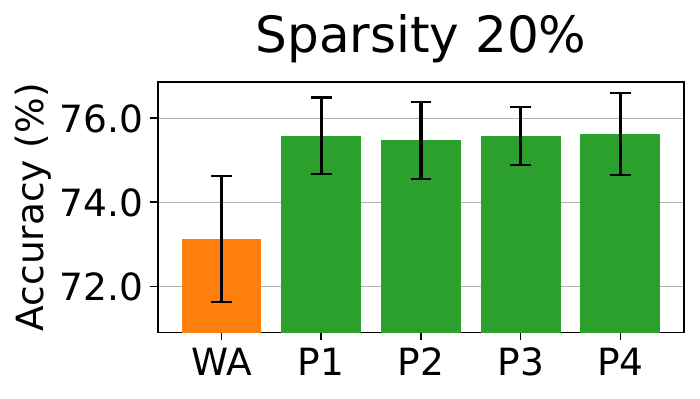}
\includegraphics[width=0.3\linewidth]{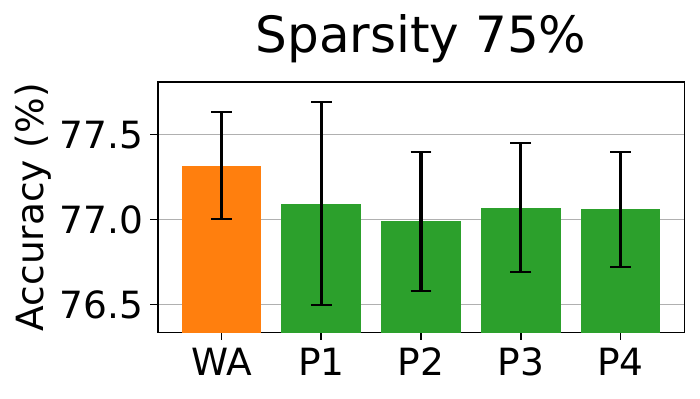}
\includegraphics[width=0.3\linewidth]{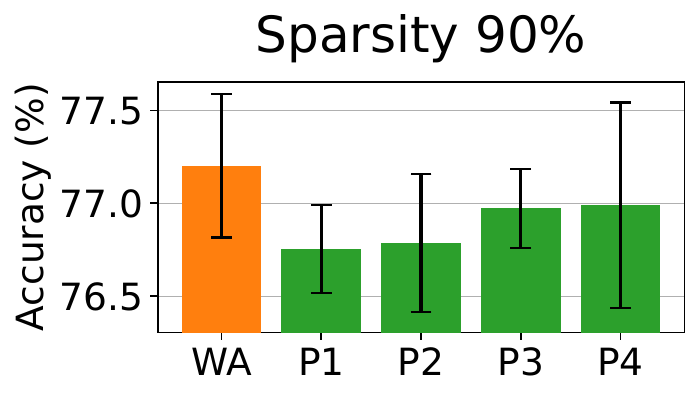}
\caption{Bar plots depicting the accuracy of individual particles involved in the averaging process of the SWAMP algorithm. While the averaged weight (denoted as WA) may not outperform individual particles (denoted as P1-P4) in the early stages of IMP (left; Sparsity 20\%), it achieves high performance at high sparsity levels. The results are presented for WRN-32-4 on the test split of CIFAR-100.}
\label{figure/particle_appendix}
\end{figure}

We provide the particle-wise accuracies for CIFAR-100 in \cref{figure/particle_appendix}. Individual SWAMP particles outperform the weight-averaged particle in lower sparsities, but we can observe the opposite trend in higher sparsities which is the area of our particular interest. 

\subsection{Connectivity between successive SWAMP solutions}
\label{app:sec:additional_experiments:connectivity}

\begin{figure}[t]
    \centering
    \includegraphics[width=\linewidth]{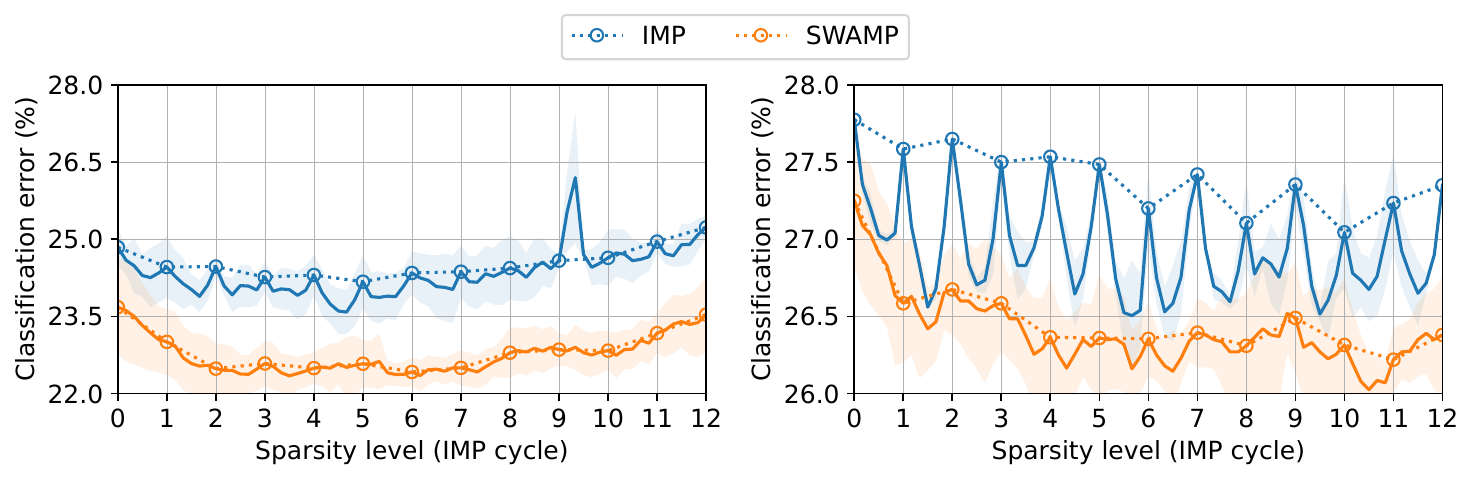}
    \caption{Linear connectivity between sparse solutions having different sparsity along with the IMP cycle. The results are presented for WRN-32-4 (left) and VGG-16 on the test split of CIFAR-100.}
    \label{figure/loss_barrier_appendix}
\end{figure}

We present additional results on CIFAR-100 regarding the linear connectivity between two consecutive SWAMP solutions. Following \cite{paul2023unmasking}, we find that SWAMP solutions from different cycles are also linearly well-connected.

\end{document}